\documentclass[times,twocolumn,final]{elsarticle}
\usepackage{amssymb}
\usepackage{latexsym}

\usepackage{url}
\usepackage{xcolor}

\usepackage{natbib}
\usepackage{hyperref}

\usepackage{amsmath,amssymb,amsfonts}
\usepackage{algorithmic}
\usepackage{graphicx}
\usepackage{textcomp}

\usepackage{makecell}
\usepackage{bm}
\usepackage{array}
\usepackage{threeparttable}
\usepackage{color}
\definecolor{R}{rgb}{1,0,0}
\usepackage{booktabs,diagbox,float}

\usepackage{algorithm}
\usepackage{algorithmic}

\definecolor{newcolor}{rgb}{.8,.349,.1}

\journal{ }

\usepackage{amsmath}
\usepackage{array}
\usepackage{booktabs}
\usepackage{multirow}
\usepackage{color}
\usepackage{colortbl}
\definecolor{R}{rgb}{1,0,0}
\definecolor{B}{rgb}{0,0,1}
\definecolor{g}{gray}{0.95}
\definecolor{G}{rgb}{0,0.8,0}
\definecolor{O}{rgb}{1,0.7,0}
\usepackage{overpic}

\begin{document}

\begin{frontmatter}

\title{ModeTv2: GPU-accelerated Motion Decomposition Transformer for\\Pairwise Optimization in Medical Image Registration}

\author[]{Haiqiao Wang}
\author[]{Zhuoyuan Wang}
\author[]{Dong Ni}
\author[]{Yi Wang\corref{cor1}}
\cortext[cor1]{
	Corresponding author: Yi Wang (onewang@szu.edu.cn)}
\address[]{School of Biomedical Engineering, Shenzhen University Medical School, Shenzhen University, Shenzhen, China}
\address[]{Smart Medical Imaging, Learning and Engineering (SMILE) Lab, Shenzhen University, Shenzhen, China}
\address[]{Medical UltraSound Image Computing (MUSIC) Lab, Shenzhen University, Shenzhen, China}

\begin{abstract}
Deformable image registration plays a crucial role in medical imaging, aiding in disease diagnosis and image-guided interventions.
Traditional iterative methods are slow, while deep learning (DL) accelerates solutions but faces usability and precision challenges.
This study introduces a pyramid network with the enhanced motion decomposition Transformer (ModeTv2) operator, showcasing superior pairwise optimization (PO) akin to traditional methods.
We re-implement ModeT operator with CUDA extensions to enhance its computational efficiency.
We further propose RegHead module which refines deformation fields, improves the realism of deformation and reduces parameters.
By adopting the PO, the proposed network balances accuracy, efficiency, and generalizability.
Extensive experiments on three public brain MRI datasets and one abdominal CT dataset demonstrate the network's suitability for PO, providing a DL model with enhanced usability and interpretability.
\emph{The code is publicly available at \url{https://github.com/ZAX130/ModeTv2}.}
\end{abstract}

\end{frontmatter}


\section{Introduction}
\begin{figure}[t]
	\centering
	\includegraphics[width=1\columnwidth]{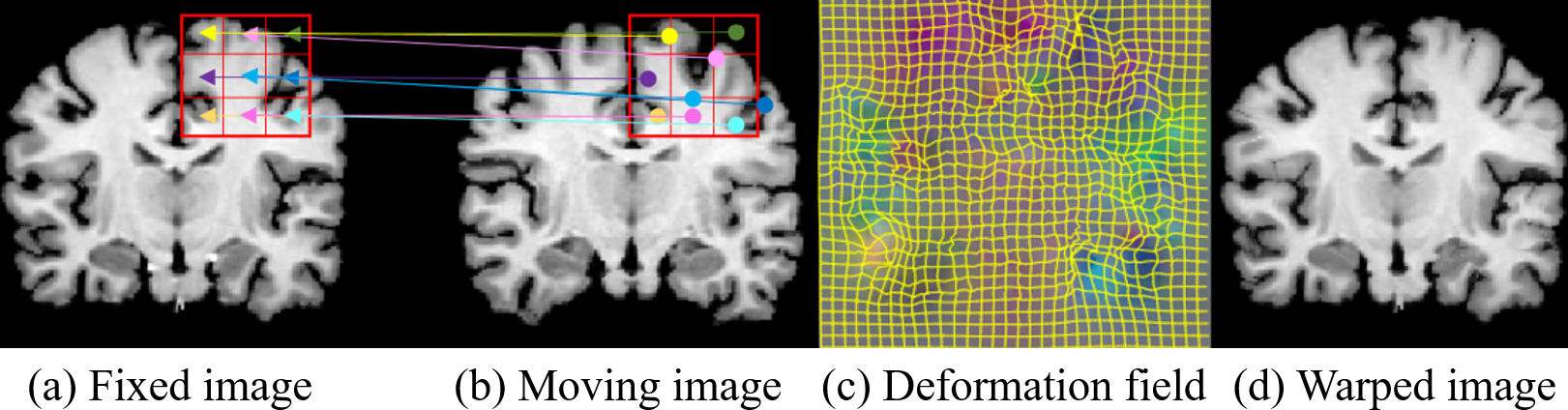}
	\caption{Illustration of the deformable image registration. Given a pair of fixed image (a) and moving image (b), the deformable registration is to estimate a non-rigid deformation field (c) to warp the moving image (d) to match with the fixed one.}
	\label{fig:intro}
\end{figure}

Deformable image registration (DIR) has always been an important focus in the society of medical imaging, which is widely used for the preoperative planning, disease diagnosis and  intraoperative guidance~\citep{sotiras2013deformable, ferrante2017slice, fu2020deep}.
The deformable registration is to solve the non-rigid deformation field to warp the moving image, so that the warped image can be anatomically similar to the fixed image, as shown in Fig.~\ref{fig:intro}.
By doing so, information from mono-/multi-modalities can be fused into the same coordinate system to assist doctors with diagnosis and treatment~\citep{ungi2016open, 9815265, song2022cross, rao2022salient}.
Although currently there are many DIR methods available, registration accuracy and efficiency are still challenges.

Traditional medical image registration methods~\citep{Bajcsy1989, Thirion1998, Rueckert1999, LDDMM, Rueckert2006, AVANTS2008, Vercauteren2009, HEINRICH201657} consider registration as an iterative process, gradually achieving well-registered images by iteratively optimizing the similarity between a pair of images.
We refer to this registration process as pairwise optimization (PO).
Such conventional approaches often require a substantial number of iterations to obtain a reasonable deformation field, which is evidently time-consuming.
However, they possess the advantage of direct applicability to different domains (i.e., different anatomical structures and image modalities),
unlike current deep learning methods that require retraining on different domains.

Deep learning (DL) approaches treat registration as a learning task~\citep{fu2020deep, haskins2020deep}, wherein these methods extensively train on a training set to obtain a registration model.
Subsequently, upon obtaining the model, there is no need for iterative computing on a new pair of images; instead, the deformation field is directly predicted through one forward pass.
Although DL methods largely accelerate the solving speed of deformation fields, there still remains challenges.
Firstly, in comparison to traditional methods, the usability of DL methods are constrained to the specific scenario presented in their training domain.
Re-training a network for a new domain entails extra data and time.
Secondly, even for the same domain task, the accuracy and interpretability of existing DL methods could be further improved.

Our motivation is directed toward identifying a DL model that not only balances accuracy and efficiency but also exhibits generalizability and interpretability.
We aspire for this model to execute PO registration akin to traditional methods.
Our optimism is grounded in the intention to enhance the inductive bias of DL registration models for the registration task.
This involves adopting network designs closely aligned with the computational aspects of the registration task to alleviate the learning burden on the network and enhance convergence efficiency, thereby reducing the pressure of retraining.

To this end, on one hand, progressively optimizing the deformation field from coarse to fine is a commonly employed strategy in DL registration tasks~\citep{Mok2020, Dual, Lv2022, Chen2022, ma2023pivit, zheng2024residual, 10423043}.
The underlying assumption is that the network can first address global deformation at low resolution and then handle local deformation at high resolution.
Such assumption exhibits a strong inductive bias tailored to the registration task, a strategy also adopted by traditional methods.
Models employing this strategy can converge to good results with fewer training epochs due to the strong inductive bias inherent in the coarse-to-fine optimization assumption.
However, most current pyramid networks consume substantial computational resources.
Additionally, these DL methods neglect multiple motion patterns at low-resolution voxels, which may lead to cumulative registration errors~\citep{zheng2024residual}.
In contrast, some traditional methods~\citep{papiez2014implicit, vishnevskiy2016isotropic, hua2017multiresolution} have explored various types of motion characteristics.

On the other hand, with the popularity of Transformer~\citep{dosovitskiy2020vit} structures, some recent registration studies have employed cross-attention mechanisms~\citep{Song2021, shi2022xmorpher, khor2023anatomically, zhu2023similarity} to capture spatial relationships between moving and fixed image features, aiming to enhance the model's inductive bias for registration.
Specifically, in these methods, one set (either the moving or fixed image) is used as the Query set, while the other set provides the Key and Value sets.
The correlation between the Query and Key sets is computed to weight the values in the Value set.
This step, akin to finding one-to-one correspondences in registration, is recognized as a promising strategy.
However, most current methods utilize above mechanism merely to enhance feature representation, with few considering its direct usage for solving the deformation field~\citep{Liu2022, Chen2022}.

In our previous work~\citep{wang2023modet}, we propose a novel motion decomposition Transformer (ModeT) for deformable image registration.
This operator simultaneously considers the decomposition of motion patterns in features and explicitly calculates the deformation field from the correlations of features, showcasing interpretability.
In this study, we further improve the registration method of our previous one~\citep{wang2023modet} and explore its capability for pairwise optimization in image registration.
Specifically, we re-implement ModeT operator with CUDA extensions to enhance its computational efficiency.
Moreover, we design a RegHead module to refine the deformation field.
The efficacy of the proposed method is evaluated on three public brain MRI datasets and one abdominal CT dataset.
Our method provides a DL model with enhanced accuracy and usability.
Our contributions can be summarized as follows:
\begin{itemize}
\item[$\bullet$] We provide a CUDA implementation of the attention computation module in ModeT to effectively improve running speed and reduce memory usage, facilitating pairwise optimization and enhancing usability.
\item[$\bullet$] We introduce a simple RegHead module to generate refined deformation fields, increasing deformation realism and reducing parameters.
\item[$\bullet$] Extensive experiments demonstrate that our method outperforms current state-of-the-art DL registration methods. By adopting the PO, the proposed method balances accuracy, efficiency, and generalizability.
\end{itemize}

\section{Related Work}
\subsection{Traditional Registration Methods}
Traditional registration methods involve iteratively optimizing a constrained deformation model to maximize similarity between fixed and moving images.
Various nonlinear deformation methods have been proposed,
such as optical flow based Demons~\citep{Thirion1998},
B-spline constrained free-form deformation (FFD)~\citep{Rueckert1999},
large deformation diffeomorphic metric mapping (LDDMM)~\citep{LDDMM},
and standard symmetric normalization (SyN)~\citep{AVANTS2008}, etc.
Commonly used similarity descriptors for measuring image similarity during registration include sum of squared differences (SSD)~\citep{Wolberg2000},
normalized mutual information (NMI)~\citep{Knops2006},
normalized cross-correlation (NCC)~\citep{Yoo2009},
modality independent neighborhood descriptor (MIND)~\citep{Heinrich2012},
and self-similarity context (SSC)~\citep{heinrich2013towards}.

The primary limitation of these methods lies in their lengthy processing time, especially when dealing with volumetric data.
Although some methods attempt to improve the computational efficiency, the processing time can still be improved.
For example, the study~\citep{heinrich2013towards} leverages discretized optimization framework, coarse grid spacing, and quantization to accelerate processing, but still takes approximately 20 seconds to conduct multimodal registration.
Note that the advantage of most traditional methods lies in the direct applicability to any pair of images without the need for pre-training, providing convenient to use.
For instance, tools like the ANTs~\citep{avants2009advanced} library,
which integrates SyN~\citep{AVANTS2008},
and the Elastix~\citep{klein2009elastix} registration package,
which integrates FFD registration~\citep{Rueckert1999},
allow users to simply input the images to directly perform registration without concerning themselves with image anatomy or modality.
Another advantage of some traditional registration methods is that they have explored various types of motion characteristics.
\cite{papiez2014implicit} replace Gaussian smoothing with bilateral filtering, which considers both spatial smoothness and local intensity similarity to preserve discontinuities in the deformation field.
\cite{vishnevskiy2016isotropic} propose the use of isotropic total variation (TV) regularization to accurately capture non-smooth displacement fields.
\cite{hua2017multiresolution} incorporate discontinuities by enriching B-spline basis functions with additional degrees of freedom.

\subsection{Structures of Deep Registration}
\subsubsection{Single Stage}
Single-stage DL registration networks are among the earliest to emerge.
The most well-known single-stage network is VoxelMorph~\citep{VoxelMorph}, which employs a Unet-like structure~\citep{Ronneberger2015} to regress the deformation field.
This structure is easy to be implemented, but it is not good enough to model complex displacements.

\subsubsection{Cascading and Recurrent Structure}
To address the issue of large deformations, especially in the context of unsupervised registration tasks, some studies utilize cascading or recurrent neural networks.
The recursive cascaded networks (RCN)~\citep{Zhao2019} concatenates multiple networks to handle the large deformation problem.
Later, the recurrent registration neural networks (R2N2)~\citep{sandkuhler2019recurrent} and the recurrent registration network (RRN)~\citep{9721919} use gated recurrent units (GRU)~\citep{69e088c8129341ac89810907fe6b1bfe} and convolutional long short term memory (CLSTM)~\citep{shi2015convolutional} modules, respectively, to construct recurrent networks for continuous deformation.
The idea behind cascading and recurrent methods is to further optimize the registration results generated by the preceding networks.
The advantage is to handle large deformation, but these methods are computationally demanding and involve redundant feature extraction.

\subsubsection{Pyramid Structure}
To alleviate the computational burden of cascading and recurrent methods, pyramid registration methods have been investigated.
The pyramid registration aims to address large deformation at low resolution first and then handle local deformation at high resolution.
Currently, there are two types of pyramid registration models.
One is similar to cascading networks, but the input at each level is a pyramid of the image itself, solving the deformation field from coarse to fine,
as described in~\citep{Mok2020, hering2019mlvirnet}.
The other type constructs the deep feature pyramid and employs these features entirely for pyramid registration~\citep{hu2019dual, Cao2021, Dual, Lv2022, Liu2022, meng2022non, ma2023pivit}.
These models usually consist of dual encoders and several flow estimators.
Specifically, the deep feature maps of the moving and fixed images are first obtained, and these features serve as inputs to the flow estimators, progressively obtaining the deformation field from coarse to fine resolution.
Additionally, some studies combine pyramid structures with recurrent networks, such as~\citep{RDN, zhou2023self}.
The main drawback of pyramid structures is that they are prone to accumulating registration errors, as each voxel at low resolution covers a large portion of the original image.
If there is a registration error at low resolution, it will propagate level-by-level and may become unrecoverable at high resolution.
Some studies design different components to address this issue, which will be discussed in the next subsection.

\subsection{Components of Deep Registration}
\subsubsection{Convolution}
Early registration methods utilize convolutional neural networks (CNNs) to construct models~\citep{VoxelMorph, Zhao2019, hu2019dual}.
With the further development of CNNs, some advanced visual modules have been continuously incorporated into registration models.
Encouraged by optical flow estimation, \cite{Dual} introduce a correlation layer to enhance the correlation estimation of the registration model.
\cite{zheng2024residual} enhance the separability of the deformation field by adding dilated convolutions~\citep{Chen2018}.
However, the addition of dilated convolutions slows down the computation, especially at high-resolution layers.
Some studies~\citep{li2022dual} employ channel and spatial attention modules~\citep{hu2018squeeze, woo2018cbam} to perform convolutional attention calculations to re-weight feature maps.
\cite{Lv2022} design specific attention modules to refine feature maps and deformation fields, respectively.

\subsubsection{Self-Attention}
The vision Transformer (ViT)~\citep{dosovitskiy2020vit} structure has been employed in registration models to consider long-range dependencies and capture richer features.
\cite{Chen2021} replace the bottom layer of the Unet structure with ViT modules.
Furthermore, \cite{Chen2022a} use the SwinEncoder~\citep{liu2021Swin} instead of the Unet encoder for feature enhancement.
Compared to the ViT structure, Swin adopts a relative friendly computing method with a moving window, leading to a more reasonable computational load.
Some studies~\citep{li2022dual, Zhu2022} also adopt similar designs.
Recently, \cite{meng2023non} and \cite{ma2023pivit} propose pyramid structures and employ SwinTransformer at each level in the decoder as a powerful flow field estimator.

\begin{figure*}[t]
	\centering
	\includegraphics[width=0.87\textwidth]{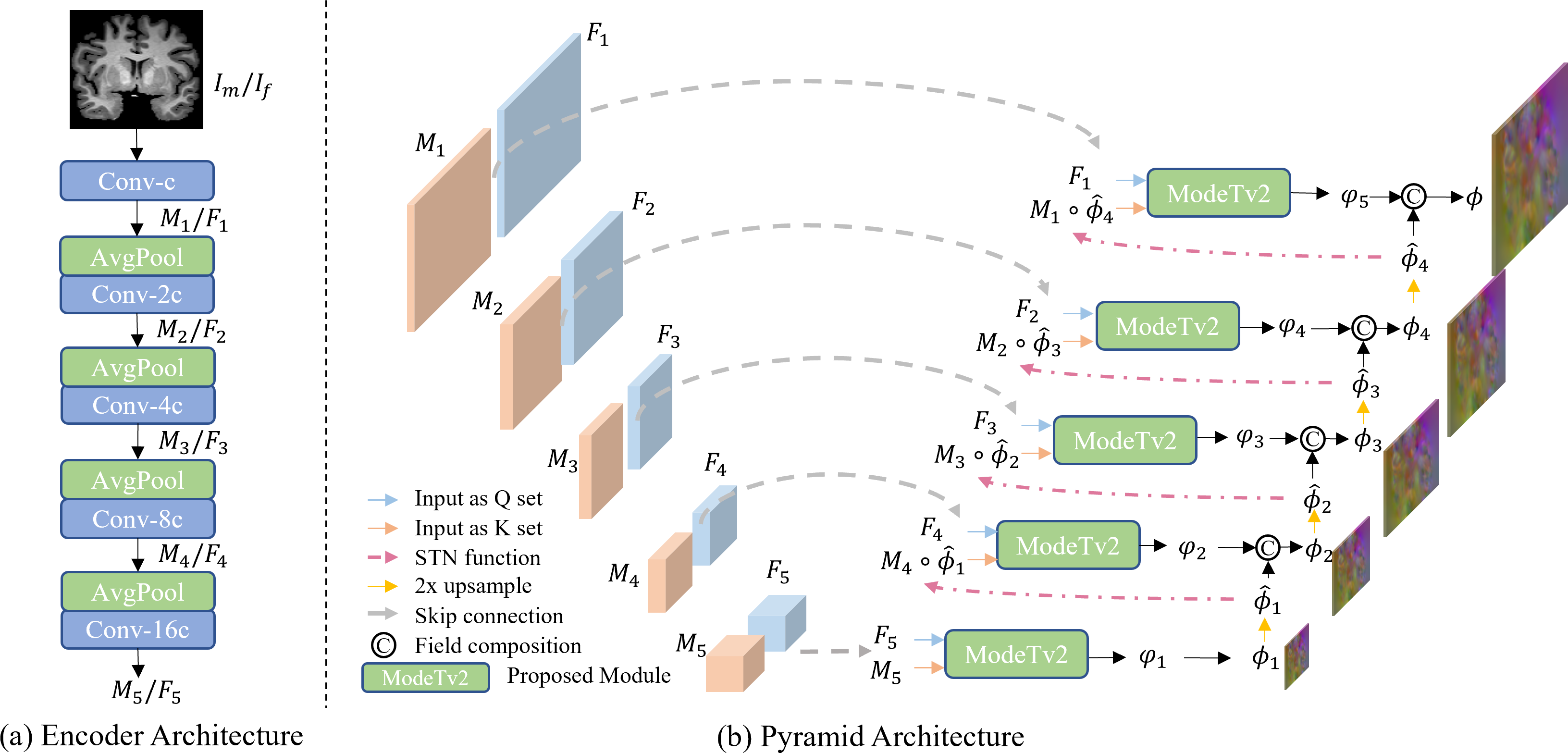}
	\caption{Illustration of the proposed deformable registration network.
		The encoder takes the fixed image $I_f$ and moving image $I_m$ as input to extract hierarchical features $F_1$-$F_5$ and $M_1$-$M_5$.
		The ModeTv2 consists of a GPU-accelerated motion decomposition Transformer (ModeT) and a registration head (RegHead).
		The ModeTv2 is used to generate multiple deformation subfields and then fuses them. Finally the decoding pyramid outputs the total deformation field $\phi$.}
	\label{fig:framework}
\end{figure*}

\subsubsection{Cross-Attention}
Furthermore, some studies have begun to consider using cross-attention~\citep{vaswani2017attention} to capture relationships between features from moving and fixed images.
Several studies~\citep{chen2023deformable, zheng2022recursive, chen2023transmatch} use cross-attention encoders to enhance the feature extraction.
\cite{Song2021} employ two cross-attention modules at the bottom layer of the U-shaped structure to perform bidirectional attention calculations on feature maps from two modalities to capture cross-modal features.
\cite{shi2022xmorpher} use cross-attention at each layer from the encoder to the decoder.
Aforementioned methods essentially employ the attention mechanisms to enhance feature representation.
Only few researches directly solve the deformation field through attention calculations; however, this is a strong inductive bias for the registration task.
\cite{Liu2022} calculate the matching scores of the feature maps from moving and fixed images.
The computed scores are then employed to re-weight the deformation grid for obtaining the final deformation field.
\cite{Chen2022} utilize the multiplication of the attention map and Value matrix in the Transformer to weight the predicted bases to generate the deformation field.
However, its attention map calculation merely concatenates and projects the moving and fixed feature maps without employing a similarity computation module.
In our previous study, we propose the ModeT~\citep{wang2023modet} operator, which uses multi-head neighborhood attention to weight the deformation grid, obtaining multiple motion modes at each level.
This allows direct calculation of the deformation field while addressing the issue of error propagation in registration.
However, the ModeT does not consider neighborhood relationships when fusing multiple motion modes, and also it could be implemented in a more efficient way.

\subsection{Generalization Studies in Deep Registration}
Due to the nature of unsupervised learning, registration networks are allowed to continue unsupervised optimization on new data, aiming to bring registered images closer in terms of similarity~\citep{VoxelMorph}.
\cite{ferrante2018adaptability} specifically investigate the pairwise optimization capability of a single-stage CNN network, with the goal of exploring whether DL model can conduct registration in a way similar to traditional methods.
The results show continuous PO on the same domain can further improve accuracy.
However, for registration models transferred from a completely different domain (different modality or organ), even when trained to convergence, they still fail to match the performance of models directly trained on the target domain.
Subsequent studies~\citep{fechter2020one, mok2023deformable} adopt similar PO strategies.
However, these methods generally focus on addressing the domain shift issue and do not discuss the handling capacity for entirely different domains.
Furthermore, advanced network designs such as attention mechanisms have not been considered.

This paper primarily focuses on exploring the PO capability of the proposed ModeTv2.
We aim to obtain a DL model with strong PO capability relying on its own structure.
Specifically, we hope that the model is with favorable accuracy, efficiency and generalizability.

\begin{figure*}[t]
	\centering
	\includegraphics[width=0.78\textwidth]{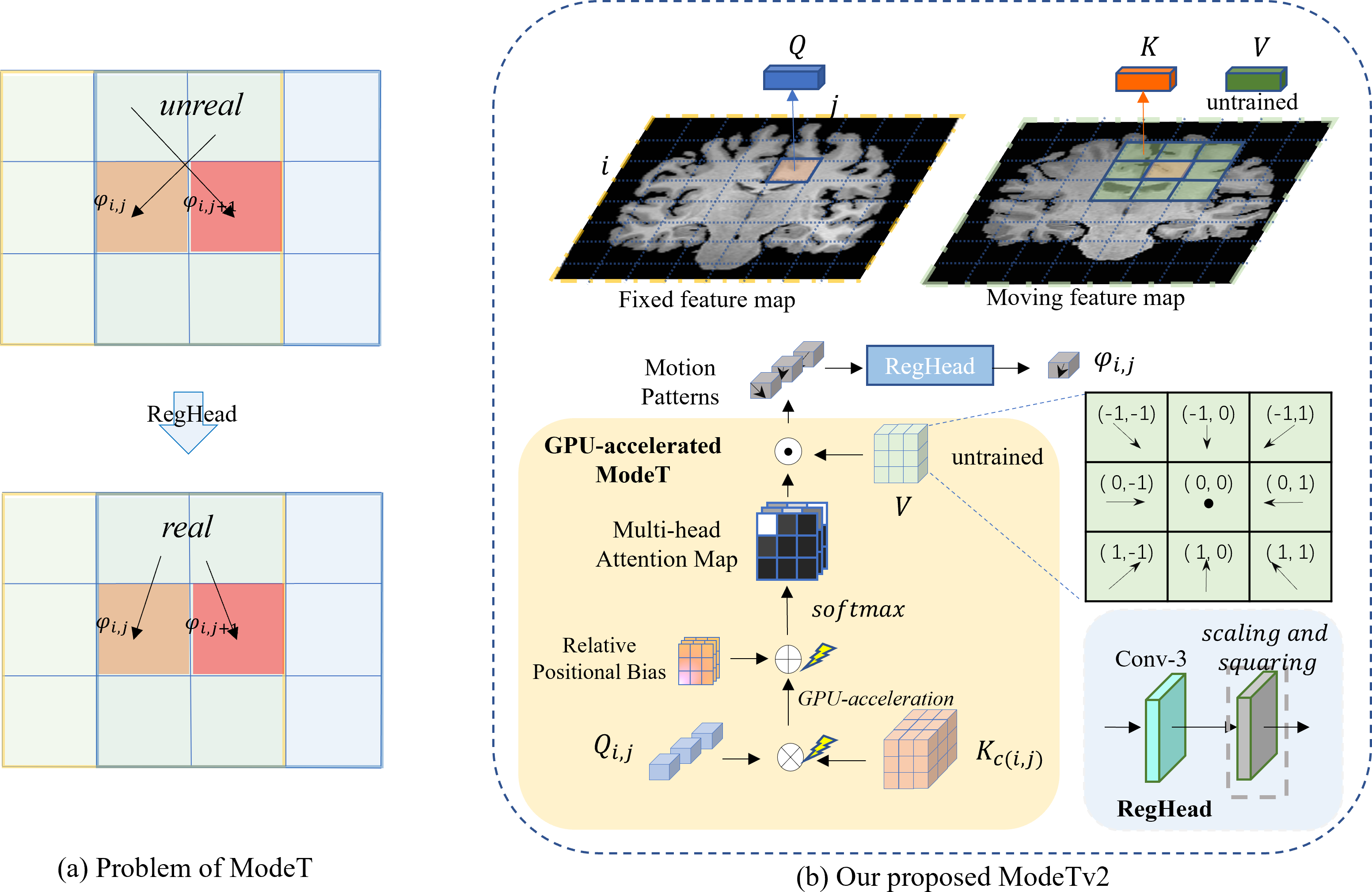}
	\caption{Illustration of the computation process for the residual subfield \(\varphi_p\) at position \(p=(i,j)\).
	Subfigure (a) illustrates the potential limitation of our previous ModeT,
	while subfigure (b) showcases the computation process of ModeTv2.
	(For the ease of understanding, we show the operations in 2D here, and $S=3$ in this illustration)	
	}
	\label{fig:ModeT}
\end{figure*}

\section{Method}
\subsection{Network Overview}
The proposed deformable registration network is illustrated in Fig.~\ref{fig:framework}.
We employ a pyramidal registration structure, which has the advantage of reducing the scope of attention calculation required at each decoding level and therefore alleviating the computational consumption.
Initially, two sets of feature maps, \(F_1, F_2, F_3, F_4, F_5\), and \(M_1, M_2, M_3, M_4, M_5\), are generated for the fixed image $I_f$ and moving image $I_m$, respectively.
Subsequently, the ModeTv2 is utilized as the deformation field estimator for each level to obtain the residual deformation field \(\varphi_l ~ (l = 1,2,3,4,5)\) at each level.
Specifically, as shown in Fig.~\ref{fig:framework}(b), the feature maps \(M_5\) and \(F_5\) are input into the ModeTv2, resulting in the coarsest level residual deformation field, denoted as \(\varphi_1\) , serving as the initial value for the total deformation field \( \phi_1 \).
The deformed moving feature map $M_4$, using the upsampled \( \hat{\phi}_1 \), is input into the next-level ModeTv2 along with the fixed feature map $F_4$ to obtain \(\varphi_2\).
The composition of \(\varphi_2\) with the previous upsampled total deformation field \( \hat{\phi}_1 \) yields the updated total deformation field \( \phi_2 \), employing the composition formula from~\citep{zhao2019unsupervised}.
Similar operations are performed for feature maps \(M_3, M_2, M_1\), and \(F_3, F_2, F_1\).
Ultimately, utilizing the obtained total deformation field \( \phi \), the image \( I_m \) is warped to produce the registered image.

To guide the network training, the normalized cross correlation $\mathcal{L}_{\text{ncc}}$~\citep{rao2014application} and the deformation regularization $\mathcal{L}_{\text{reg}}$~\citep{VoxelMorph} is used:
\begin{equation}
	\theta^*=\underset{\theta}{argmin}\mathcal{L}_{\text{ncc}}(I_f, I_m\circ\phi_\theta) + \lambda \mathcal{L}_{\text{reg}}(\phi_\theta),
	\label{e:loss}
\end{equation}
where $\circ$ is the warping operation,
$\lambda$ is the weight of the regularization term,
\( \theta \) represents the network parameters,
and \( \phi_\theta \) denotes the deformation field generated by the registration network.
The equation implies that through continuous optimization of \( \theta \), the registration results improve progressively.
During the network training phase, \( I_m \) and \( I_f \) are two distinct images selected from the training set, and the equation aims to achieve a globally optimal solution.
In the PO scenario, \( I_m \) and \( I_f \) represent a specific image pair, and the equation seeks to obtain the optimal solution \( \theta^* \) for this particular pair of images.

\subsection{Encoder}
The encoder in the pyramid registration network is employed to extract features for subsequent matching.
It is desirable that this component remains relatively uncomplicated to prevent feature misalignment within the encoder.
Therefore, we adopt a concise design for the encoder.

As illustrated in Fig.~\ref{fig:framework}(a), our encoder consists of five hierarchical levels \((L = 1,2,3,4,5)\) to generate hierarchical representations of moving and fixed features.
The encoders for encoding fixed and moving feature maps share weights, facilitating the generation of features suitable for matching.
In Fig.~\ref{fig:framework}(a), $Conv$ represents consecutive pairs of 3$\times$3$\times$3 convolutions, instance normalization, and leaky ReLU activation, with the number of channels doubling at each layer.
Simultaneously, average pooling layers are utilized at each level for downsampling the features.

\subsection{Motion Decomposition Transformer Version 2 (ModeTv2)}
\subsubsection{GPU-accelerated ModeT}
\textbf{Design of ModeT:}
In DL registration networks, each position \( p \) in the low-resolution feature map captures rich semantic information from a sizable portion of the original image, allowing for diverse potential motion modes.
To tackle this, we leverage a sophisticated multi-head neighborhood attention mechanism as shown in Fig.~\ref{fig:ModeT}(b), to dissect various motion modes at the low-resolution level.

Let \( F, M\in \mathbb{R}^{c\times h\times w\times l} \) be the fixed and moving feature maps from a specific layer of the hierarchical encoder, where \( h, w, l \) are feature map dimensions, and \( c \) is the channel count.
These feature maps undergo linear projection (\( proj \)) and LayerNorm (\( LN \)) operations~\citep{Ba2016layer} to generate \( Q \) (query) and \( K \) (key):
\begin{equation}
	\begin{split}
		Q = LN(&proj(F)), \quad K= LN(proj(M)), \\
		Q=&\{Q^{(1)}, Q^{(2)},\dots,Q^{(S)}\},\\
		K=&\{K^{(1)}, K^{(2)},\dots,K^{(S)}\},
	\end{split}
\end{equation}
where the shared-weight projection operation originates from \( N(0,1e^{-5}) \), with biases initialized to 0.
\( Q \) and \( K \) are then divided based on channels,
resulting in \( Q, K\in \mathbb{R}^{S\times h\times w\times l\times d} \), where \( S \) represents the number of attention heads, and \( d \) is the number of channels per attention head.

The neighborhood attention map is then calculated.
Let \( c(p) \) be the voxel \( p \)'s neighborhood,
where \( p\in \mathbb{R}^{3} \), representing the position of any voxel in \( Q \).
For a neighborhood of size \( n\times n\times n \), \( ||c(p)||=n^3 \).
The multi-head neighborhood attention map is obtained through:
\begin{equation}
	NA(p,s) = softmax(Q_p^{(s)}\cdot K^{(s)T}_{c(p)}+B^{(s)}), 
	\label{e:na}
\end{equation}
where \( B\in\mathbb{R}^{S\times n\times n \times n} \) is a learnable relative positional bias, initialized to zeros;
and \( NA(p,s)\in \mathbb{R}^{n\times n\times n} \).
Zero padding is applied to the moving feature map to handle boundary voxels.
Equation~(\ref{e:na}) shows how the neighborhood attention is computed for the \( s \)-th head at position \( p \), breaking down semantic information from low resolution for individual similarity computation, laying the foundation for modeling diverse motion modes. 

Subsequently, multiple subfields at this level are obtained by computing the regular displacement field, weighted via the neighborhood attention map:
\begin{equation}
	\varphi_p^{(s)} = NA(p,s)V,
	\label{e:subsubflow}
\end{equation}
where \( \varphi^{(s)}\in\mathbb{R}^{h\times w \times l\times 3} \),
\( V\in\mathbb{R}^{n\times n \times n \times 3} \),
and \( V \) (value) represents the relative position coordinates for the neighborhood centroid.
Specifically, $V_x = x - x_c$, where \( x\in \mathbb{R}^{3} \) denotes a voxel's position of \( V \), and $x_c = \left( \lceil n/2 \rceil, \lceil n/2 \rceil, \lceil n/2 \rceil \right)$ is the central voxel's position of \( V \).
Importantly, \( V \) remains unlearned, allowing the multi-head attention relationship to naturally transform into a multi-coordinate association.

These steps result in a series of motion subfields at this tier:
\begin{equation}
	\varphi^{(1)},\varphi^{(2)},\dots,\varphi^{(S)}
	\label{e:warps}
\end{equation}
It's worth noting that as decoding feature maps become more refined, the number of motion modes at position \( p \) decreases, along with the required number of attention heads $S$. 

\textbf{CUDA Implementation:}
We utilize CUDA extensions to re-implement the computation of ModeT, aiming to enhance both training and inference efficiency.
Specifically, in mainstream DL frameworks such as PyTorch or TensorFlow, the dot product operation on the neighborhood in ModeT can only be performed by first creating a sliding window and then computing the dot product.
This computation method leads to excessive GPU memory consumption due to the sliding window, and introduces additional computational overhead due to step-wise calculations.

To address these challenges, we employ CUDA extensions to skip the step of creating a sliding window, allowing for an overhead similar to convolution.
This optimization effectively reduces GPU memory usage and minimizes additional computational costs associated with step-wise calculations.

\subsubsection{Registration Head (RegHead)}
We reevaluate the challenges associated with the diverse motion modes obtained through ModeT.
In~\citep{wang2023modet}, a competitive weighting module (CWM) is designed to select a predominant motion pattern from various motion modes at a single point.
However, weighting alone is insufficient to adequately address issues such as local field crossings that arise from multiple subfields (see Fig.~\ref{fig:ModeT}(a)), since ModeT does not account for neighborhood relationships during calculation, or more precisely, only considers point to-point relationships.
Simultaneously, to achieve a highly generalized network, we aim to minimize the number of trainable parameters on the deformation field estimator.
We prefer the network to focus more on relearning the encoder's feature extraction during PO.

For the sake of convenience in transferability and efficiency, we propose a registration head (RegHead) module.
As illustrated in Fig.~\ref{fig:ModeT}(b), the RegHead employs a single layer of ~\(3\times3\times3\) convolution to integrate multiple motion modes and results in $\varphi_l$ at each layer $l$.
The weights of the convolution layer are initialized to \(N(0, 1e^{-5})\).
Importantly, we continue to use RegHead even when there is only one predominant motion mode at high resolutions, ensuring the reasonability of deformation fields at each layer.
Furthermore, to preserve the topological properties of deformation fields, we introduce an optional diffeomorphic layer within RegHead.

Under the condition of employing the diffeomorphic layer, we define the output of the convolution layer in RegHead as the stationary velocity field \(v\)~\citep{dalca2019unsupervised}.
With regard to time $t$, the deformation field $\phi$ is defined as the ordinary differential equation (ODE):
\begin{equation}
\frac{d\phi^{(t)}}{dt}=v(\phi^{(t)}), 
\label{e:ode}
\end{equation}
where $\phi^{(t)}$ is an identity transformation when $t =0$, and the residual deformation field at this level when $t=1$.
We integrate $\phi^{(t)}$ from $t=0$ to $t=1$ by employing the scaling and squaring (ss) operation~\citep{arsigny2006log}.
Specifically, $\phi^{(1)}$ is achieved by repeating the loop:
$\phi^{(\frac{1}{2^{t+1}})}= \phi^{(\frac{1}{2^{t}})}\circ \phi^{(\frac{1}{2^{t}})}$.
After $T$ iterations, we obtain the final deformation field $\varphi$ at this level.

It is worth noting that RegHead acts as a filter for motion patterns generated by ModeT.
The convolutional layers within RegHead can learn kernels resembling mean or Gaussian filters to resolve most crossing artifacts.
Furthermore, the ss layer refines the deformation field by performing multiple-step integration of velocity fields.
This effectively decomposes large deformations into small diffeomorphic steps, preserving diffeomorphism over iterations.

\section{Experiments}
\subsection{Datasets}
To verify the efficacy of the proposed method, experiments were carried on four public datasets, including three brain MRI datasets and one abdomen CT dataset.

\textbf{LPBA dataset}~\citep{Shattuck2008} contains 40 brain MRI volumes, and each with 54 manually labeled region-of-interests (ROIs).
All volumes have been rigidly pre-aligned to mni305.
30 volumes (30$\times$29 pairs) were employed for training and 10 volumes (10$\times$9 pairs) were used for testing.

\textbf{Mindboggle dataset}~\citep{Klein2012} has been affinely aligned to mni152.
Each brain MRI volume contains 62 manually labeled ROIs.
42 volumes (42$\times$41 pairs from the NKI-RS-22 and NKI-TRT-20 subsets) were employed for training, and 19 volumes from MMRR-21 (19$\times$18 pairs) were used for testing.
All MRI volumes from the LPBA and Mindboggle datasets were pre-processed by min-max normalization, and skull-stripping using FreeSurfer~\citep{Fischl2012}.
The final size of each volume was 160$\times$192$\times$160 (1mm$\times$1mm$\times$1mm) after a center-cropping operation.

\textbf{IXI dataset}\footnote{\url{https://brain-development.org/ixi-dataset/}},
pre-processed by~\citep{Chen2022a},
includes brain MRI scans that have undergone affine pre-registration to the Talairach space, intensity normalization, and center cropping to dimensions of 160$\times$192$\times$224 (1mm$\times$1mm$\times$1mm).
Each data contains 30 labeled ROIs.
Following the protocol of~\citep{Chen2022a}, we conducted template-to-subject registration, utilizing 403 samples for training and 115 for testing.
This task is typically employed for atlas-based segmentation.

\textbf{Abdomen CT (ABCT) dataset}\footnote{\url{https://learn2reg.grand-challenge.org/}} has undergone canonical affine pre-alignment. 
Each CT volume contains 4 annotated organs, including liver, spleen, left kidney, and right kidney.
All CT volumes were center-cropped to a size of 160$\times$160$\times$192 (2mm$\times$2mm$\times$2mm) by utilizing masks provided by the dataset.
35 volumes (35$\times$34 pairs) were used for training,
and 8 volumes (8$\times$7 pairs) were used for testing.   

\subsection{Evaluation Metrics}
\textbf{Accuracy evaluation.}
Dice score (DSC)~\citep{Dice1945} was computed as the primary similarity metric to evaluate the degree of overlap between corresponding regions.
Moreover, the average symmetric surface distance (ASSD)~\citep{Taha2015} and the 95\% maximum Hausdorff distance (HD95)~\citep{Huttenlocher1993} were calculated, which can reflect the similarity of the region contours.
The quality of the predicted non-rigid deformation $\phi$ was assessed by the percentage of voxels with non-positive Jacobian determinant (i.e., folded voxels).
All above metrics were calculated in 3D.

\textbf{Efficiency evaluation.}
The running time, GPU memory and trainable parameter number required for the volumetric registration were recorded.

An ideal registration shall have higher DSC as well as lower HD95, ASSD, and Jacobian values within a shorter running time, less GPU memory and less trainable parameter number.

\subsection{Comparison Methods}
We compared our method with several cutting-edge registration methods:
(1)~\texttt{SyN}~\citep{AVANTS2008}: a classical traditional approach, using the $SyNOnly$ setting in ANTS.
(2)~\texttt{VoxelMorph(VM)}~\citep{VoxelMorph}: a popular CNN-based single-stage registration network.
(3)~\texttt{VoxelMorph-diff(VM-diff)}~\citep{dalca2019unsupervised}: a popular CNN-based single-stage registration network.
(4)~\texttt{TransMorph(TM)}~\citep{Chen2022a}: a single-stage registration network with SwinTransformer-enhanced encoder.
(5)~\texttt{TransMatch}~\citep{chen2023transmatch}: a Transformer-based registration network using cross attention modules for each level of the encoder and decoder.
(6)~\texttt{PR++}~\citep{Dual}: a pyramid registration network using 3D correlation layer.
(7)~\texttt{DeFormer}~\citep{Chen2022}: a pyramid registration network using Deformer module and multi-resolution refinement module.
(8)~\texttt{PCnet}~\citep{Lv2022}: a pyramid network using several attention and weighting strategies to fuse multi-level feature maps and deformation fields.
(9)~\texttt{PIViT}~\citep{ma2023pivit}: a light pyramid network using iterative SwinTransformer blocks in low resolution.
(10)~\texttt{Im2grid(I2G)}~\citep{Liu2022}: a pyramid registration network with the coordinate Transformer.
(11)~\texttt{ModeT}~\citep{wang2023modet}: our conference version.

\subsection{Implementation Details}
We tested four different versions of our method, namely ModeTv2-s, ModeTv2-l, ModeTv2-diff-s, and ModeTv2-diff-l,
which are the small model, large model, and their corresponding diffeomorphism versions with ss operation, respectively.
The small model had $c$=8 convolutional channels in the first layer, as well as 6 channels for each head in the multi-head attention mechanism.
The number of attention heads were set as 8, 4, 2, 1, 1 from bottom to top, respectively.
This setting is the same as our conference version ModeT~\citep{wang2023modet}, allowing for easy comparison.
Due to the implementation of CUDA operators, the large model version can be trained.
Specifically, the large model had $c$=32 convolutional channels in the first layers, as well as 12 dimensions for each attention head.
In the pyramid decoder, from coarse to fine, the number of attention heads were set as 32, 16, 8, 4, 1, respectively.

\textbf{Training details on source domain.}
Our method was implemented with PyTorch, using a GPU of NVIDIA Tesla V100 with 32GB memories.
The CUDA version we used is 11.3.

Firstly, for VM-diff, \(\sigma\) was set to 0.01, and \(\lambda\) was set to 20 across all datasets.
Then, regarding the remaining DL networks,
the values of \(\lambda\) were configured as follows:
for the single-stage network on the LPBA dataset, \(\lambda\) was set to 4,
while for the pyramid network, it was set to 1.
On the Mindboggle, ABCT, and IXI datasets, all networks were assigned \(\lambda\) values of 1, 0.5, and 4, respectively.
For SyN, we used mean square error (MSE) as loss function.
The neighborhood size $n$ was set as $3$.
We used the Adam optimizer~\citep{kingma2014adam} with a learning rate decay strategy as follows:
\begin{equation}
	lr_m =  lr_{init}\cdot(1-\frac{m-1}{M})^{0.9}, m = 1, 2, ... ,M
\end{equation}
where $lr_m$ represents the learning rate of $m$-th epoch and $lr_{init}$ represents the learning rate of initial epoch.
For training the single-stage network on the LPBA dataset, we employed  $lr_{init}=0.0004$.
For the IXI dataset, $lr_{init}=0.0002$ was used.
For all other training tasks, $lr_{init}=0.0001$ was applied.
We set the batch size as $1$, $M$ as $30$ for the training of pyramid methods and $50$ for single-stage methods.

\begin{table}[t]
	\centering
	\caption{The numerical results of different registration methods on the ABCT (4 ROIs) dataset.}
	\label{tab:abct}
	\scriptsize
	\begin{tabular}{p{0.18\textwidth}p{0.04\textwidth}p{0.05\textwidth}p{0.05\textwidth}c}
		\toprule
		Methods & DSC ($\%$) & HD95 (mm)& ASSD (mm)& $ \% |J_{\phi}|\le0 $\\
		\cmidrule(l{2pt}r{2pt}){1-5}
		\texttt{Initial} & 45.7$\pm$10.3 & 33.53$\pm$9.89 & 12.26$\pm$4.15 &-  \\
		\texttt{SyN}~\citep{AVANTS2008} & 48.7$\pm$14.2 & 35.05$\pm$12.87 & 12.43$\pm$5.65 & $ <$0.2\%  \\
		\texttt{VM}~\citep{VoxelMorph} & 58.5$\pm$15.5 & 32.77$\pm$11.97 & 10.45$\pm$5.29 & $ <$11\%  \\
		\texttt{VM-diff}~\citep{dalca2019unsupervised} & 51.2$\pm$14.3 & 35.32$\pm$13.43 & 12.13$\pm$5.63 & $ <$1\%  \\
		\texttt{TM}~\citep{Chen2022a} & 63.9$\pm$16.5 & 31.09$\pm$12.19 & 9.26$\pm$5.37 & $ <$14\%  \\
		\texttt{TransMatch}~\citep{chen2023transmatch} & 63.7$\pm$16.4 & 31.37$\pm$12.33 & 9.29$\pm$5.37 & $ <$13\%  \\
		\texttt{PR++}~\citep{Dual} & 68.8$\pm$15.9 & 29.22$\pm$12.98 & 8.08$\pm$5.26 & $ <$6\%  \\
		\texttt{DeFormer}~\citep{Chen2022} & 64.6$\pm$16.0 & 32.34$\pm$12.73 & 9.23$\pm$5.34 & $ <$12\%  \\
		\texttt{PCnet}~\citep{Lv2022} & 72.4$\pm$15.5 & 27.00$\pm$12.91 & 7.10$\pm$5.10 & $ <$0.2\%  \\
		\texttt{PIViT}~\citep{ma2023pivit} & 71.2$\pm$14.1 & 26.92$\pm$12.19 & 7.07$\pm$4.08 & $ <$2\%  \\
		\texttt{I2G}~\citep{Liu2022} & 72.3$\pm$14.5 & 27.38$\pm$12.93 & 6.91$\pm$4.58 & $ <$0.9\%  \\
		\texttt{ModeT}~\citep{wang2023modet} & 77.4$\pm$12.6 & 24.07$\pm$11.80 & 5.50$\pm$3.67 & $ <$3\%  \\
		\texttt{ModeTv2-s} & 78.5$\pm$12.7 & 23.61$\pm$12.47 & 5.44$\pm$3.83 & $ <$2\%  \\
		\texttt{ModeTv2-diff-s} & 77.8$\pm$13.4 & 23.85$\pm$12.85 & 5.59$\pm$4.04 & $ <$0.009\%  \\
		\texttt{ModeTv2-l} & 80.6$\pm$11.7 & 21.92$\pm$12.12 & 4.95$\pm$3.49 & $ <$4\%  \\
		\texttt{ModeTv2-diff-l} & \textbf{81.8$\pm$11.8} & \textbf{19.57$\pm$11.67} & \textbf{4.48$\pm$3.39} & $ <$0.2\%  \\
		\bottomrule
	\end{tabular}
\end{table}

\begin{figure}[t]
	\centering
	\includegraphics[width=1\columnwidth]{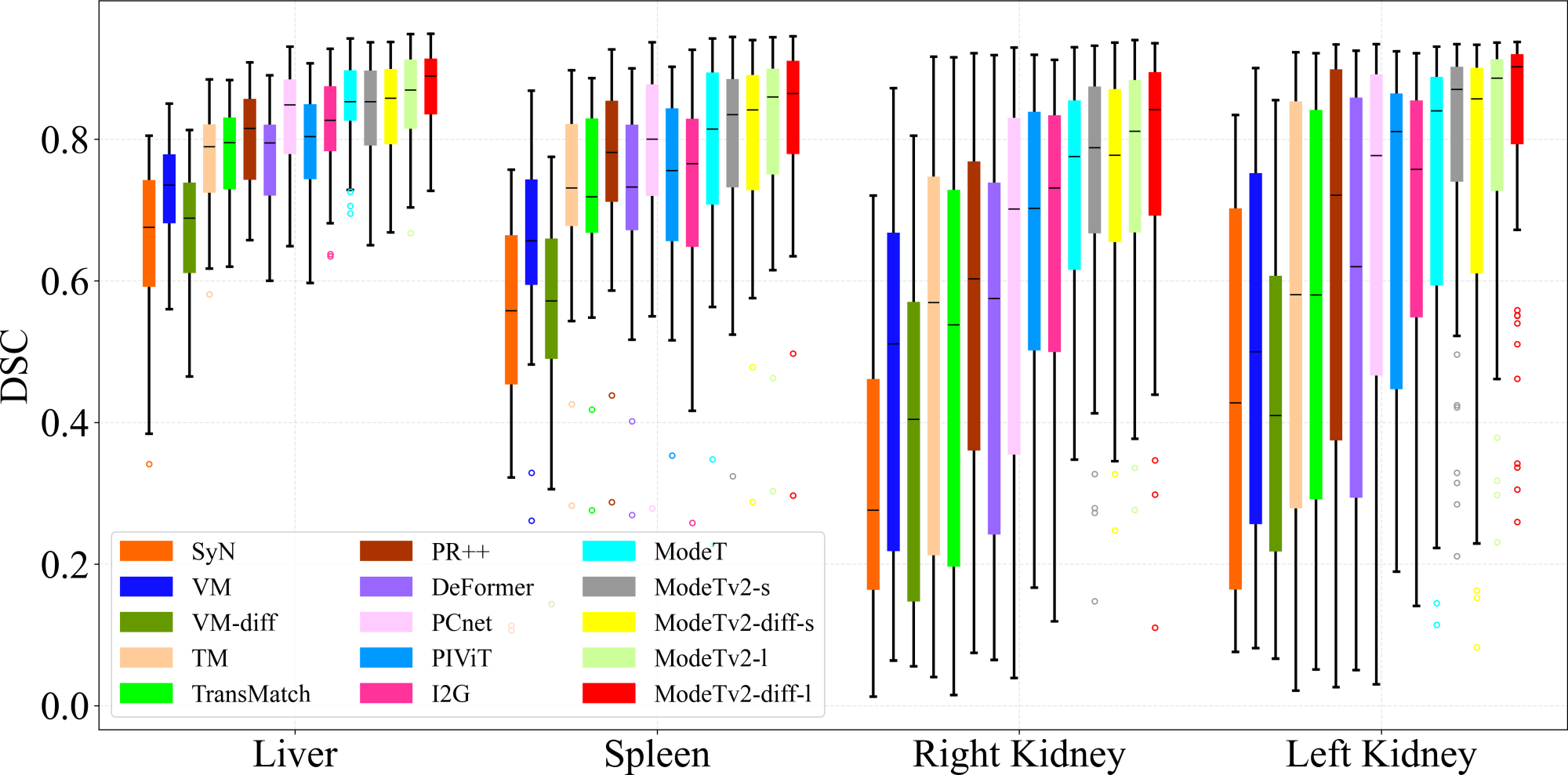}
	\caption{Box plots showing the distributions of DSC scores of four regions on the ABCT dataset produced by different registration methods.}
	\label{fig:abct}
\end{figure}

\begin{table}[t]
	\centering
	\caption{The numerical results of different registration methods on the LPBA (54 ROIs) dataset.}
	\label{tab:lpba}
	\scriptsize
	\begin{tabular}{p{0.18\textwidth}p{0.035\textwidth}p{0.04\textwidth}p{0.04\textwidth}c}
		\toprule
		Methods & DSC ($\%$) & HD95 (mm)& ASSD (mm)& $ \% |J_{\phi}|\le0 $\\
		\cmidrule(l{2pt}r{2pt}){1-5}
		\texttt{Initial} & 53.7$\pm$4.8 & 7.43$\pm$0.78 & 2.74$\pm$0.33 &- \\
		\texttt{SyN}~\citep{AVANTS2008} & 70.4$\pm$1.8 & 5.82$\pm$0.50 & 1.72$\pm$0.12 & $ <$0.0004\%  \\
		\texttt{VM}~\citep{VoxelMorph} & 68.2$\pm$2.3 & 6.14$\pm$0.58 & 1.84$\pm$0.17 & $ <$0.004\%  \\
		\texttt{VM-diff}~\citep{dalca2019unsupervised} & 65.6$\pm$2.8 & 6.39$\pm$0.62 & 1.97$\pm$0.19 & $ <$0.0001\%  \\
		\texttt{TM}~\citep{Chen2022a} & 68.9$\pm$2.4 & 6.12$\pm$0.62 & 1.82$\pm$0.18 & $ <$0.01\%  \\
		\texttt{TransMatch}~\citep{chen2023transmatch} & 67.8$\pm$2.5 & 6.20$\pm$0.61 & 1.87$\pm$0.18 & $ <$0.009\%  \\
		\texttt{PR++}~\citep{Dual} & 69.5$\pm$2.2 & 6.11$\pm$0.59 & 1.76$\pm$0.17 & $ <$0.2\%  \\
		\texttt{DeFormer}~\citep{Chen2022} & 69.2$\pm$2.4 & 6.14$\pm$0.62 & 1.79$\pm$0.18 & $ <$0.4\%  \\
		\texttt{PCnet}~\citep{Lv2022} & 72.0$\pm$1.9 & 5.71$\pm$0.55 & 1.62$\pm$0.14 & $ <$0.00003\%  \\
		\texttt{PIViT}~\citep{ma2023pivit} & 70.8$\pm$1.5 & 5.67$\pm$0.48 & 1.67$\pm$0.11 & $ <$0.04\%  \\
		\texttt{I2G}~\citep{Liu2022} & 71.0$\pm$1.4 & 5.69$\pm$0.47 & 1.64$\pm$0.10 & $ <$0.01\%  \\
		\texttt{ModeT}~\citep{wang2023modet} & 72.1$\pm$1.4 & 5.54$\pm$0.47 & 1.58$\pm$0.11 & $ <$0.007\%  \\
		\texttt{ModeTv2-s} & 73.0$\pm$1.3 & 5.45$\pm$0.45 & 1.53$\pm$0.10 & $ <$0.002\%  \\
		\texttt{ModeTv2-diff-s} & 73.1$\pm$1.3 & 5.46$\pm$0.45 & 1.54$\pm$0.10 & $ <$0.0000007\%  \\
		\texttt{ModeTv2-l} & \textbf{73.9$\pm$1.2} & \textbf{5.30$\pm$0.43} & \textbf{1.49$\pm$0.09} & $ <$0.02\%  \\
		\texttt{ModeTv2-diff-l} & 73.9$\pm$1.2 & 5.32$\pm$0.43 & 1.49$\pm$0.09 & $ <$0.0000003\%  \\
		\bottomrule
	\end{tabular}
\end{table}

\begin{figure}[t]
	\centering
	\includegraphics[width=1\columnwidth]{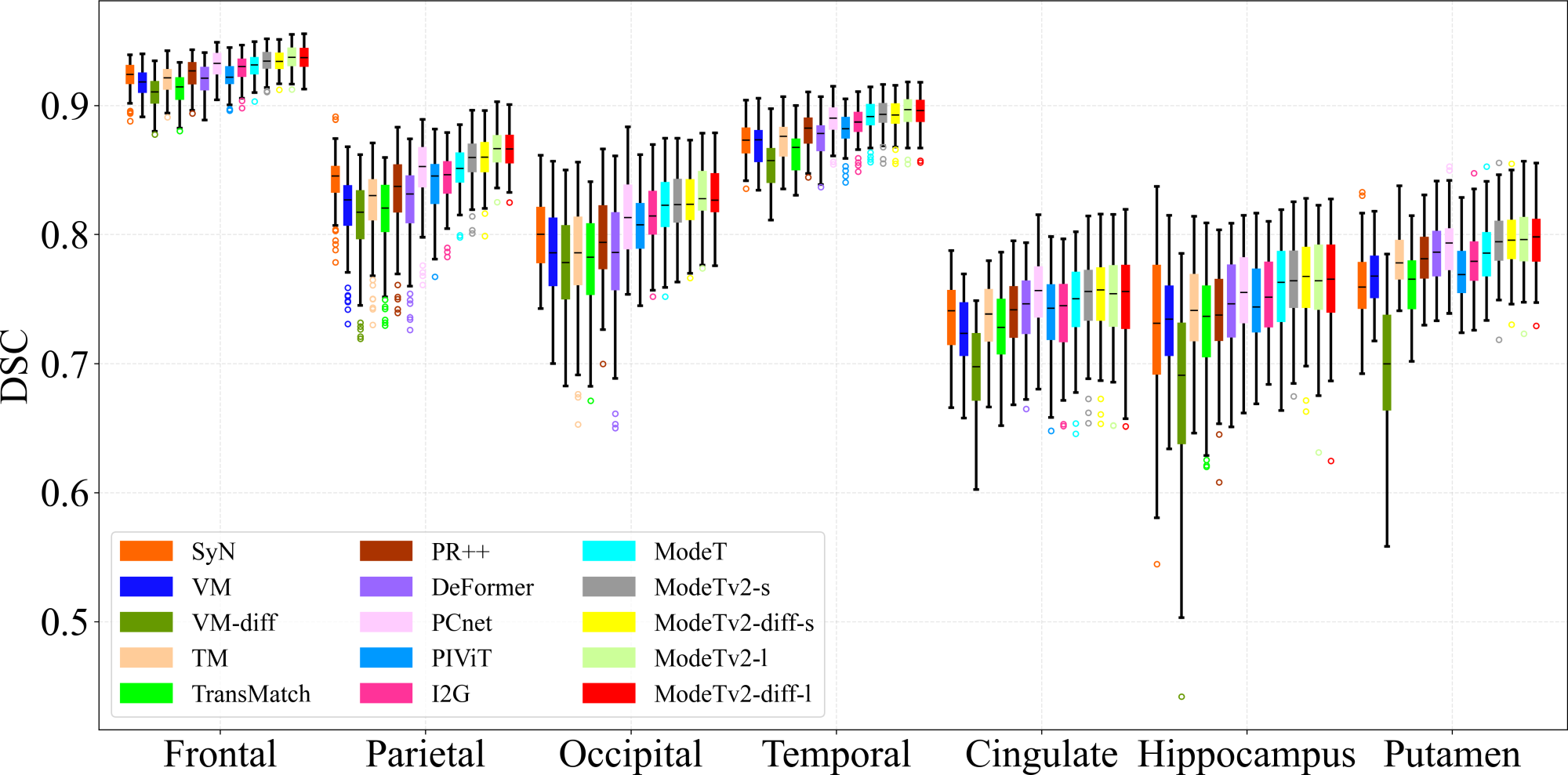}
	\caption{Box plots showing the distributions of DSC scores of seven regions on the LPBA dataset produced by different registration methods.}
	\label{fig:lpba}
\end{figure}

\textbf{PO details on target domain.}\footnote{Note that iterative optimization was used only in the PO experiments (Section~\ref{sec:po}) to evaluate the cross-domain generalization capabilities of different network structures. For the accuracy experiments (Section~\ref{sec:accuracy}) and convergence experiments (Section~\ref{sec:conv}), no iterative optimization was employed, and the methodology aligns with typical DL approaches.}
We conducted 50 iterations of fine-tuning on each model.
Note that considering IXI dataset was used for template-to-subject registration, while the other three datasets for inter-subject registration, we only employed LPBA, Mindboggle and ABCT for the PO experiments.
For fine-tuning on brain data, \(\lambda\) was set to 1, whereas abdominal data used \(\lambda=0.5\).
The learning rate was consistently maintained at $0.0001$ throughout the process.
The PO experiments were performed using the Adam optimizer,
while in a separate set of experiments, a comparison between the stochastic gradient descent (SGD) optimizer and the Adam optimizer was conducted.

The code is publicly available at \url{https://github.com/ZAX130/ModeTv2}.

\subsection{Results}
\subsubsection{Registration Accuracy}
\label{sec:accuracy}

\begin{table}[t]
	\centering
	\caption{The numerical results of different registration methods on the Mindboggle (62 ROIs) dataset.}
	\label{tab:mind}
	\scriptsize
	\begin{tabular}{p{0.18\textwidth}p{0.035\textwidth}p{0.04\textwidth}p{0.04\textwidth}c}
		\toprule
		Methods & DSC ($\%$) & HD95 (mm) & ASSD (mm)& $ \% |J_{\phi}|\le0 $\\
		\cmidrule(l{2pt}r{2pt}){1-5}
		\texttt{Initial} & 34.0$\pm$1.9 & 6.62$\pm$0.44 & 2.04$\pm$0.13 & -  \\
		\texttt{SyN}~\citep{AVANTS2008} & 53.0$\pm$2.0 & 5.65$\pm$0.52 & 1.44$\pm$0.13 & $ <$0.000006\%  \\
		\texttt{VM}~\citep{VoxelMorph} & 54.7$\pm$1.7 & 6.07$\pm$0.47 & 1.50$\pm$0.12 & $ <$1\%  \\
		\texttt{VM-diff}~\citep{dalca2019unsupervised} & 51.2$\pm$2.5 & 6.02$\pm$0.46 & 1.54$\pm$0.12 & $ <$0.009\%  \\
		\texttt{TM}~\citep{Chen2022a} & 58.5$\pm$1.7 & 5.83$\pm$0.50 & 1.38$\pm$0.13 & $ <$0.9\%  \\
		\texttt{TransMatch}~\citep{chen2023transmatch} & 56.0$\pm$1.7 & 5.84$\pm$0.50 & 1.42$\pm$0.13 & $ <$1\%  \\
		\texttt{PR++}~\citep{Dual} & 58.4$\pm$1.6 & 5.85$\pm$0.48 & 1.39$\pm$0.12 & $ <$0.5\%  \\
		\texttt{DeFormer}~\citep{Chen2022} & 58.1$\pm$1.7 & 5.76$\pm$0.51 & 1.38$\pm$0.13 & $ <$0.7\%  \\
		\texttt{PCnet}~\citep{Lv2022} & 61.0$\pm$1.7 & 5.59$\pm$0.52 & 1.30$\pm$0.13 & $ <$0.006\%  \\
		\texttt{PIViT}~\citep{ma2023pivit}: & 56.4$\pm$1.6 & 5.48$\pm$0.50 & 1.35$\pm$0.13 & $ <$0.05\%  \\
		\texttt{I2G}~\citep{Liu2022} & 57.4$\pm$1.5 & 5.55$\pm$0.44 & 1.34$\pm$0.11 & $ <$0.02\%  \\
		\texttt{ModeT}~\citep{wang2023modet} & 60.2$\pm$1.6 & 5.38$\pm$0.46 & 1.27$\pm$0.12 & $ <$0.03\%  \\
		\texttt{ModeTv2-s} & 61.5$\pm$1.6 & 5.29$\pm$0.49 & 1.23$\pm$0.12 & $ <$0.02\%  \\
		\texttt{ModeTv2-diff-s} & 61.1$\pm$1.6 & 5.30$\pm$0.48 & 1.24$\pm$0.12 & $ <$0.0002\%  \\
		\texttt{ModeTv2-l} & 62.3$\pm$1.6 & 5.21$\pm$0.49 & 1.21$\pm$0.12 & $ <$0.08\%  \\
		\texttt{ModeTv2-diff-l} &\textbf{62.8$\pm$1.7} & \textbf{5.11$\pm$0.52} & \textbf{1.17$\pm$0.13} & $ <$0.007\%  \\
		\bottomrule
	\end{tabular}
\end{table}

\begin{figure}[t]
	\centering
	\includegraphics[width=1\columnwidth]{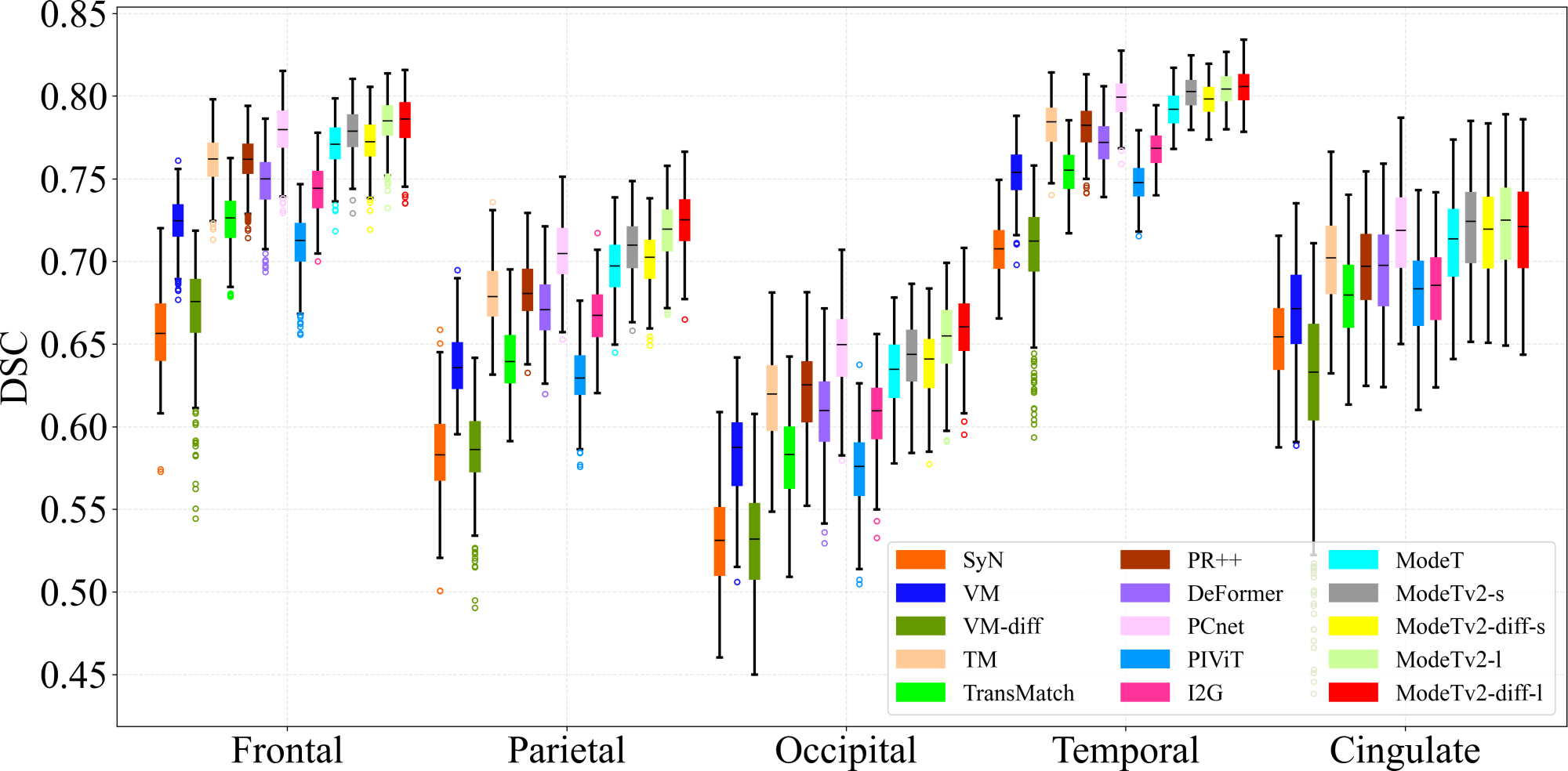}
	\caption{Box plots showing the distributions of DSC scores of five regions on the Mindboggle dataset produced by different registration methods.}
	\label{fig:mind}
\end{figure}

\begin{table}[t]
	\centering
	\caption{The numerical results of different registration methods on the IXI (30 ROIs) dataset.}
	\label{tab:ixi}
	\scriptsize
	\begin{tabular}{p{0.18\textwidth}p{0.035\textwidth}p{0.04\textwidth}p{0.04\textwidth}c}
		\toprule
		Methods & DSC ($\%$) & HD95 (mm) & ASSD (mm)& $ \% |J_{\phi}|\le0 $\\
		\cmidrule(l{2pt}r{2pt}){1-5}
		\texttt{Initial} & 40.6$\pm$3.5 & 6.24$\pm$0.63 & 2.72$\pm$0.29 & -  \\
		\texttt{SyN}~\citep{AVANTS2008} & 63.6$\pm$3.9 & 5.55$\pm$0.88 & 1.74$\pm$0.28 & $ <$0.0002\%  \\
		\texttt{VM}~\citep{VoxelMorph} & 72.4$\pm$2.5 & 2.90$\pm$0.36 & 1.08$\pm$0.12 & $ <$0.05\%  \\
		\texttt{VM-diff}~\citep{dalca2019unsupervised} & 57.5$\pm$3.6 & 5.11$\pm$0.85 & 1.85$\pm$0.25 & $ <$0.003\%  \\
		\texttt{TM}~\citep{Chen2022a} & 73.9$\pm$2.7 & 2.82$\pm$0.37 & 1.03$\pm$0.13 & $ <$0.2\%  \\
		\texttt{TransMatch}~\citep{chen2023transmatch} & 73.6$\pm$2.6 & 2.89$\pm$0.39 & 1.05$\pm$0.13 & $ <$0.2\%  \\
		\texttt{PR++}~\citep{Dual} & 74.4$\pm$2.4 & 2.78$\pm$0.37 & 1.01$\pm$0.12 & $ <$0.04\%  \\
		\texttt{DeFormer}~\citep{Chen2022} & 71.5$\pm$3.2 & 3.08$\pm$0.44 & 1.14$\pm$0.15 & $ <$0.05\%  \\
		\texttt{PCnet}~\citep{Lv2022} & 74.9$\pm$2.5 & 2.77$\pm$0.39 & 0.99$\pm$0.12 & $ <$0.00006\%  \\
		\texttt{PIViT}~\citep{ma2023pivit} & 75.2$\pm$2.1 & 2.68$\pm$0.37 & 0.97$\pm$0.10 & $ <$0.03\%  \\
		\texttt{I2G}~\citep{Liu2022} & 74.2$\pm$2.4 & 2.80$\pm$0.37 & 1.01$\pm$0.12 & $ <$0.0003\%  \\
		\texttt{ModeT}~\citep{wang2023modet} & 75.0$\pm$2.2 & 2.74$\pm$0.37 & 0.98$\pm$0.11 & $ <$0.0002\%  \\
		\texttt{ModeTv2-s} & 75.8$\pm$2.2 & 2.68$\pm$0.37 & 0.96$\pm$0.11 & $ <$0.0004\%  \\
		\texttt{ModeTv2-diff-s} & 75.7$\pm$2.2 & 2.67$\pm$0.37 & 0.96$\pm$0.11 & 0  \\
		\texttt{ModeTv2-l} & \textbf{75.9$\pm$2.2} & \textbf{2.64$\pm$0.37} & \textbf{0.95$\pm$0.11} & $ <$0.02\%  \\
		\texttt{ModeTv2-diff-l} & 75.7$\pm$2.3 & 2.69$\pm$0.39 & 0.96$\pm$0.11 & 0  \\
		\bottomrule
	\end{tabular}
\end{table}

\begin{figure}[t]
	\centering
	\includegraphics[width=1\columnwidth]{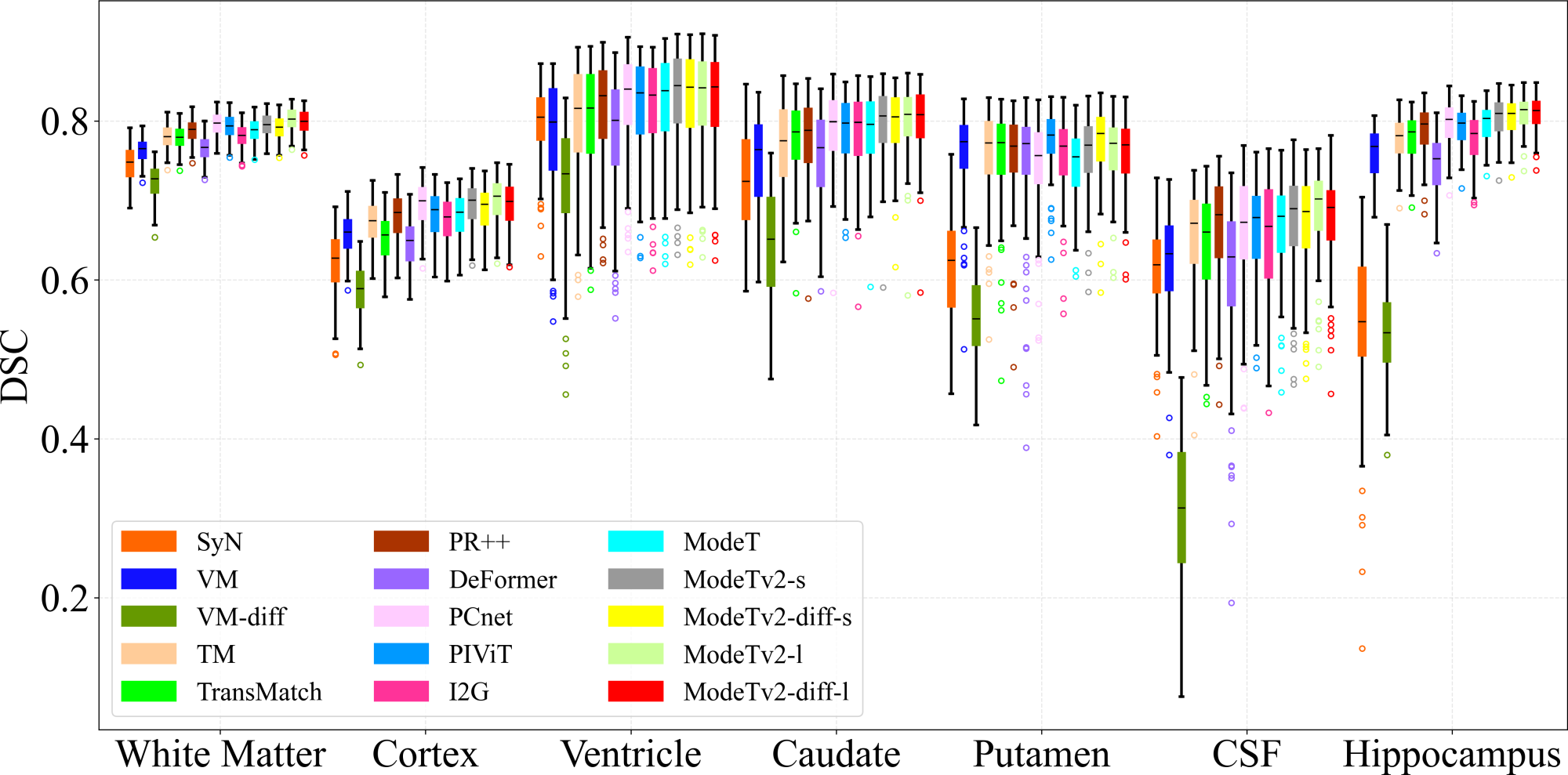}
	\caption{Box plots showing the distributions of DSC scores of seven regions on the IXI dataset produced by different registration methods.}
	\label{fig:ixi}
\end{figure}

The numerical results of different methods on datasets ABCT, LPBA, Mindboggle and IXI are listed in Tables~\ref{tab:abct},~\ref{tab:lpba},~\ref{tab:mind} and~\ref{tab:ixi}, respectively.
It can be observed that our ModeTv2, especially the large model, consistently attained satisfactory registration accuracy with respect to the metrics of DSC, HD95 and ASSD.
These Tables also report the percentage of voxels with non-positive Jacobian determinant ($ \% |J_{\phi}|\le0 $), which demonstrate our ModeTv2 predicted high quality non-rigid deformation field.
We now analyze the comparison results on each dataset.
\textbf{ABCT.}
It can be seen from Table~\ref{tab:abct} that even with affine pre-alignment, the ABCT images still had large deformation (initial DSC value of 45.7\%).
In such case, traditional SyN and single-stage networks did not perform well.
In contrast, our ModeTv2 achieved overall the best results of DSC, HD95 and ASSD.
In particular, the large version of ModeTv2 outperformed all comparison methods (even including its small counterpart) obviously.
Moreover, the diffeomorphism version of ModeTv2-s generated the lowest negative Jacobian ratio, while maintaining satisfactory registration accuracy.
\textbf{LPBA.}
As shown in Table~\ref{tab:lpba}, the small version of ModeTv2 already generated the best registration accuracy among all other comparison methods, and the large version further improved the performance.
Compared with the normal version, the diffeomorphism version of ModeTv2-l achieved almost the same registration accuracy and maintained extremely low negative Jacobian ratio.
\textbf{Mindboggle.}
The images from Mindboggle possess complex deformation and therefore the initial affine pre-alignment attained only an average DSC of 34.0\%, as listed in Table~\ref{tab:mind}.
Among all methods, our ModeTv2 again achieved favorable registration accuracy and generated high quality deformation field.
\textbf{IXI.}
The IXI dataset differs from LPBA and Mindboggle as its labeled ROIs are not predominantly concentrated in the lobar regions but are instead located near the brainstem.
Consequently, this dataset evaluates overall alignment performance rather than intricate deformation, leading to higher DSC scores for most methods.
As illustrated in Tables~\ref{tab:mind} and~\ref{tab:ixi}, PIViT, a lightweight pyramid network leveraging the Swin Transformer structure, achieved a DSC of 75.2\% on IXI, despite its moderate performance on Mindboggle.
In contrast, our model achieved the highest DSC score, with the -diff models further ensuring fold-free deformations.
It is worth noting that for all the experimental datasets, our method in this study consistently outperformed the conference version ModeT~\citep{wang2023modet},
which demonstrates the efficacy of our simple RegHead in fusion of multiple deformation subfields.
Moreover, due to the CUDA implementation, the ModeTv2 is much more time-/memory-efficient compared with ModeT, which will be discussed later.

The distributions of DSC results of specific regions on datasets ABCT, LPBA, Mindboggle and IXI are shown in Figs.~\ref{fig:abct},~\ref{fig:lpba},~\ref{fig:mind} and~\ref{fig:ixi}, respectively.
In particular, Fig.~\ref{fig:abct} illustrates the DSC distributions of four organs including liver, spleen, left kidney and right kidney;
Fig.~\ref{fig:lpba} illustrates the DSC distributions of seven anatomical regions including frontal, parietal, occipital, temporal, cingulate, hippocampus and putamen;
Fig.~\ref{fig:mind} illustrates the DSC distributions of five anatomical regions including frontal, parietal, occipital, temporal and cingulate;
Fig.~\ref{fig:ixi} presents the DSC distributions across seven anatomical regions: white matter, cortex, ventricle, caudate, putamen, cerebrospinal fluid (CSF), and hippocampus.
It can be observed that for most anatomical regions, the DSC distributions of our large ModeTv2 were positioned higher to some perceptible extent.

\begin{figure*}[t]
	\centering
	\includegraphics[width=\textwidth]{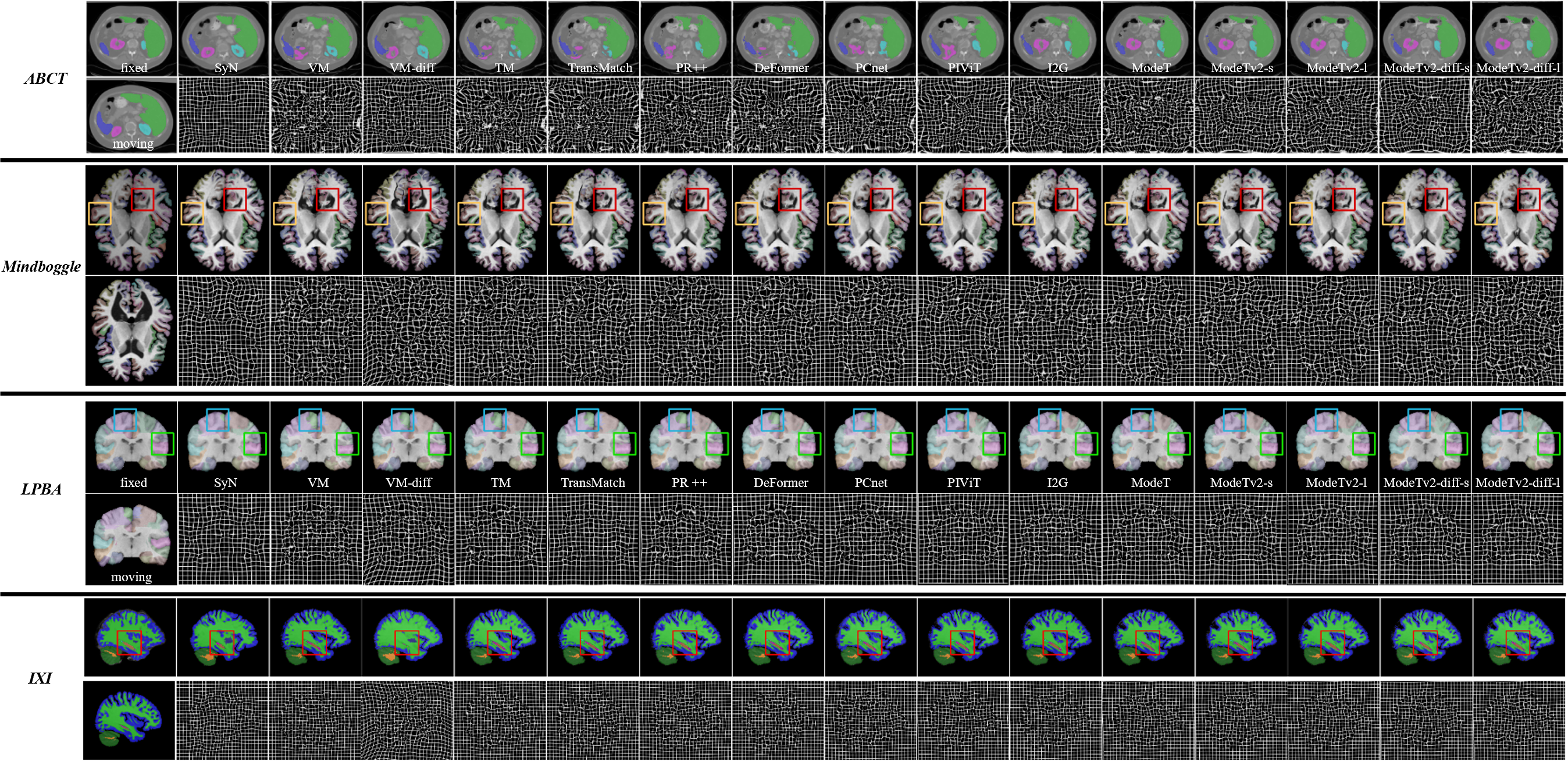}
	\caption{The visualization of the registration results and the corresponding deformation fields from different methods on four datasets.}
	\label{fig:allexamples}
\end{figure*}

\begin{figure*}[t]
	\centering
	\includegraphics[width=0.9\textwidth]{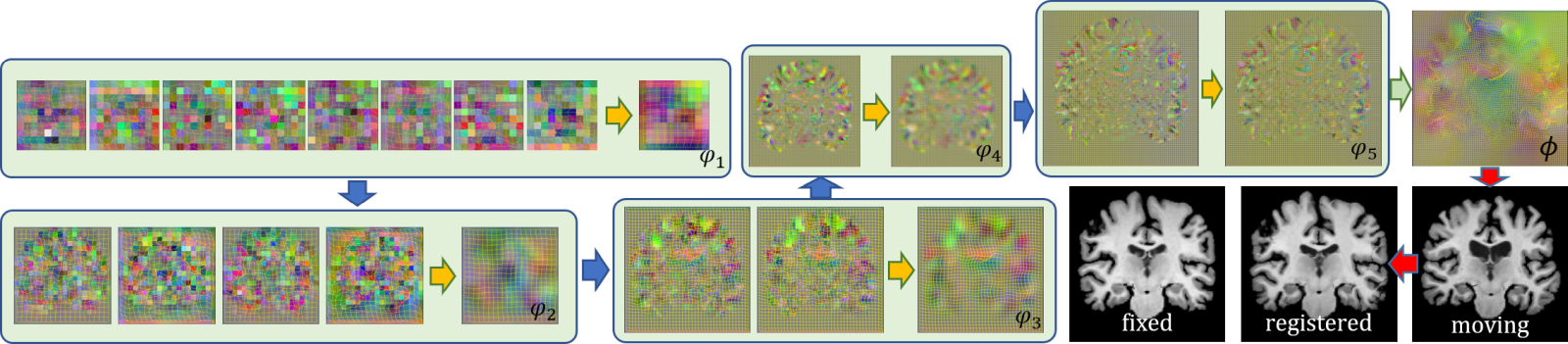}
	\caption{Visualization of the generated multi-level deformation fields ($\varphi_1$-$\varphi_5$) to register one image pair from Mindboggle. At low-resolution levels, multiple deformation subfields are decomposed to effectively model different motion modes.}
	\label{fig:deformation}
\end{figure*}

Fig.~\ref{fig:allexamples} visualizes the registered images and the corresponding deformation fields from different methods on four datasets (we encourage you to zoom in for better visualization).
Our ModeTv2 generated more accurate registered images, and the internal structures can be consistently preserved by using our method.
Moreover, our estimated deformation fields were relatively smooth, even with large displacements, the local deformations still seem reasonable.
Fig.~\ref{fig:deformation} takes the registration of one image pair as an example to show the multi-level deformation fields generated by our method using ModeTv2-diff-s.
Our ModeTv2 effectively modeled multiple motion modes and our RegHead fused them together.
The final deformation field $\phi$ accurately warped the moving image to registered with the fixed image.

\begin{figure*}[t]
	\centering
	\includegraphics[width=0.6\textwidth]{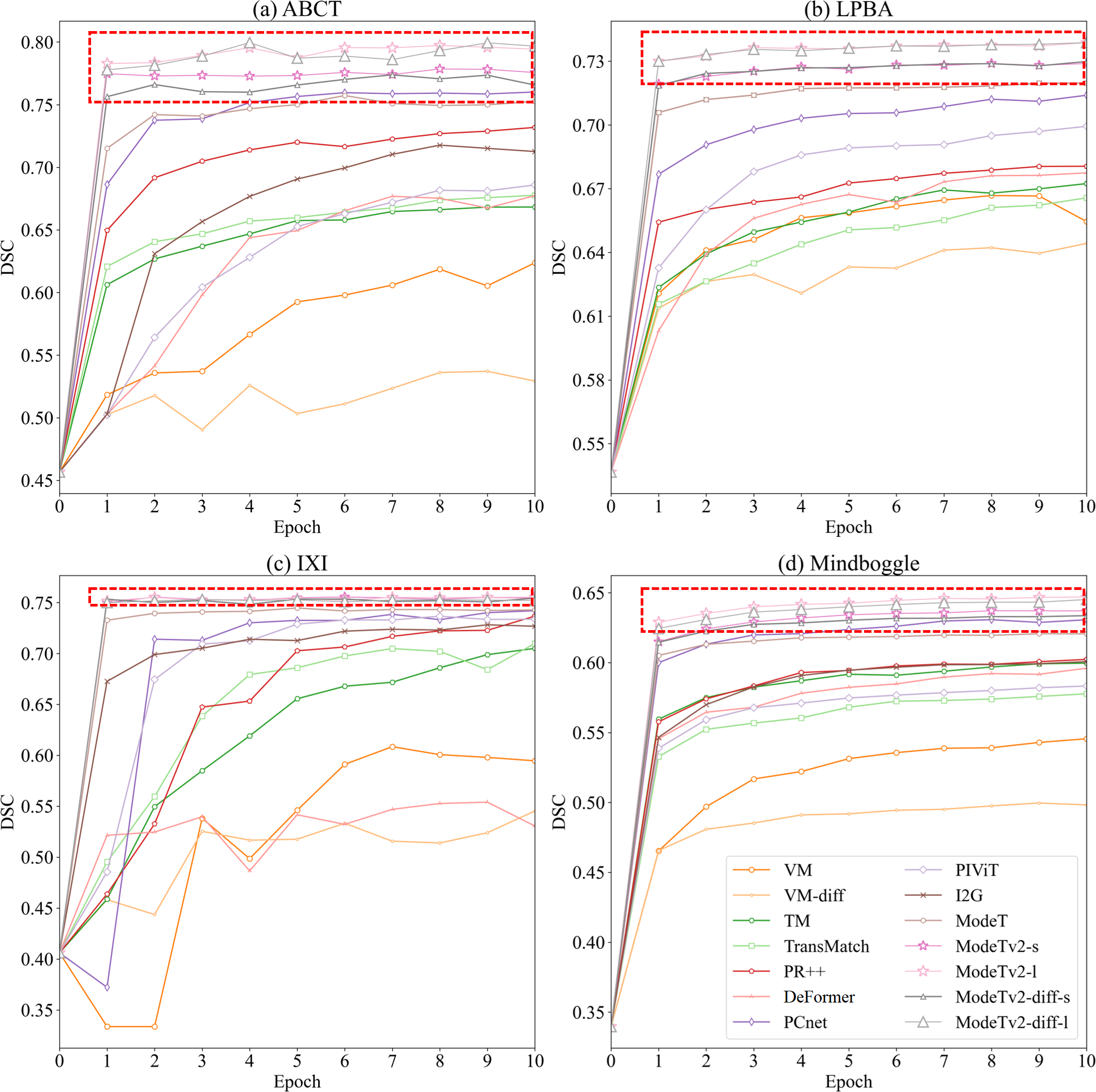}
	\caption{Illustration of the DSC results of different DL methods on four datasets during the first ten epochs of the training phase.}
	\label{fig:converge}
\end{figure*}

\subsubsection{Convergence Results}
\label{sec:conv}
Fig.~\ref{fig:converge} presents the Dice results of different DL models on the validation sets of various datasets during the first 10 epochs of the training phase.
It is evident that the single-stage CNN models (VM, VM-diff) converged slowly on these datasets.
In contrast, the single-stage models enhanced with Transformer structures demonstrate better convergence performance.
The pyramid models, overall, converged faster than the single-stage networks, with our ModeTv2 achieving the fastest convergence among these networks.

\subsubsection{PO Results}
\label{sec:po}
\begin{figure*}[t]
	\centering
	\includegraphics[width=0.75\textwidth]{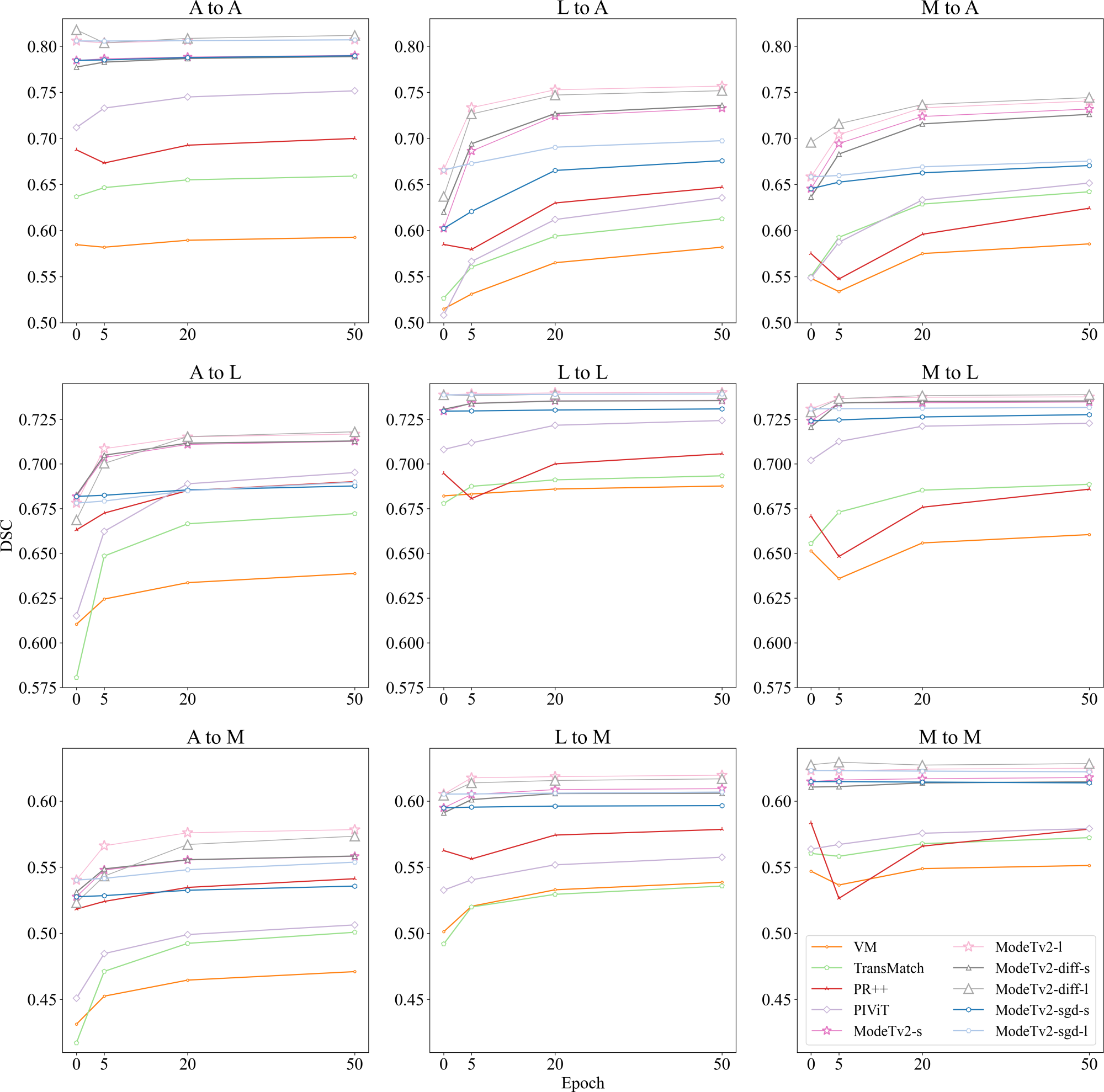}
	\caption{Illustration of same-domain and cross-domain pairwise optimization results, where A represents the ABCT dataset, L represents LPBA, M represents Mindboggle, and X to Y indicates models trained on the training set of X and tested on the test set of Y.}
	\label{fig:POres}
\end{figure*}

Fig.~\ref{fig:POres} presents the results of 9 groups of PO experiments.
In this context, we selected representative models for each architecture and compared them with our network.
Firstly, it can be observed that PO results within the same domain (A2A, L2L, M2M) were consistently higher compared to other cross-domain PO results.
Our ModeTv2-diff-l experienced a slight initial decline in the A2A experiments, but gradually recovered over subsequent stages.
In the L2L experiments, our small-scale network continued to exhibit improvement with continuous pairwise training.
In other same-domain experiments, the convergence curves of our network tended to stabilize, indicating that the network itself converged well, and further training within the same domain did not result in apparent enhancements.

Secondly, in cross-domain PO experiments across different brain MRI datasets (M2L and L2M), our model achieved relatively high DSC values even without PO (epoch=0).
Additionally, our model closely approached the corresponding models in the same-domain PO, converging within approximately 5 epochs.
For instance, ModeTv2-diff-l initiated at 72.9\% in M2L and converged to 73.9\% after 5 epochs.
PIViT exhibited faster convergence in M2L, while it converged slower in L2M, opposite to the results of PR++.
Single-stage networks showed weaker cross-domain convergence capabilities.

Furthermore, in cross-domain (different modalities and organs) PO experiments (L2A, M2A, A2M, A2L), the results of all models were lower compared to same-modal but different-center PO results.
It is noteworthy that cross-domain PO results between ABCT and LPBA were higher than those between ABCT and Mindboggle.
For M2A, except for ModeTv2-l, the initial values of our ModeTv2 models were higher than L2A, but the final convergent results were reversed.
For example, ModeTv2-diff-s achieved a result of 73.6\% after 50 epochs in L2A, while it was 72.6\% in the M2A.

Additionally, Fig.~\ref{fig:POres} also shows the PO trained using SGD optimizer for comparisons.
It can be observed that Adam was a more appropriate strategy in our PO experiments.
The SGD did not converge in many cases.

In summary, even in cross-domain PO experiments between Mindboggle and ABCT, our results compared favorably with the second-best results in same-domain PO experiments.
Our network demonstrated superior initial results and rapid convergence across all experiments, requiring no further PO in the same domain, approximately 5 epochs of PO in the same modality but different centers, and around 20 epochs of PO in cross-modal and cross-organ scenarios for convergence.

\subsubsection{Registration Efficiency}

\begin{table}[t]
	\centering
	\caption{The time and memory usage of different methods during training and inference.}
	\label{tab:spme}
	\scriptsize
	\begin{tabular}{p{0.18\textwidth}p{0.04\textwidth}p{0.05\textwidth}p{0.04\textwidth}p{0.05\textwidth}}
		\toprule
		Methods & Training Time ($ms$) &  Training GPU Mem($MB$) &  Inference Time ($ms$) &  Inference GPU Mem($MB$) \\
		\cmidrule(l{2pt}r{2pt}){1-5}
		\texttt{SyN}~\citep{AVANTS2008} & - & - & $>$10$^5$	& -   \\
		\texttt{VM}~\citep{VoxelMorph} & 1078.94 & 9174 & 102.5 & 3677  \\
		\texttt{VM-diff}~\citep{dalca2019unsupervised} & 283.73 & 4120 & 27.43 & 1974  \\
		\texttt{TM}~\citep{Chen2022a} & 1243.14 & 17906 & 201.7 & 5263 \\
		\texttt{TransMatch}~\citep{chen2023transmatch} & 1225.95 & 22922 & 240.85 & 4392 \\
		\texttt{PR++}~\citep{Dual} & 3111.99& 17584& 522.7 & 9435  \\
		\texttt{DeFormer}~\citep{Chen2022} & 3112.99 & 22979& 980.4 & 8383  \\
		\texttt{PCnet}~\citep{Lv2022}& 3436.53& 23584 & 753.87 & 6324  \\
		\texttt{PIViT}~\citep{ma2023pivit}: & 654.05 & 4506 & 28.24 & 2085  \\
		\texttt{I2G}~\citep{Liu2022} & 1195.74 & 19558 & 290.2 & 9805  \\
		\texttt{ModeT}~\citep{wang2023modet} &7595.52 &18710 &  558.3 & 9809  \\
		\texttt{ModeTv2-s} & 1069.74 & 8960 & 235.07 &  3530  \\
		\texttt{ModeTv2-diff-s} & 1166.65 & 9520 & 258.83 & 4098  \\
		\texttt{ModeTv2-l} & 1582.76 & 19808 & 277.31 & 6328  \\
		\texttt{ModeTv2-diff-l} & 1696.38 & 20450 & 301.77 &  6408  \\
		\bottomrule
	\end{tabular}
\end{table}

Table~\ref{tab:spme} shows the running time and memory usage of different methods to register and pairwise optimize on a volume pair of size 160$\times$192$\times$160.
The iterative method SyN took the longest inference time.
VM-diff exhibited the fastest runtime, followed by PIViT, which shared a similar encoder structure.
However, these two methods could not provide accurate enough registration.
PCnet generated satisfactory registration accuracy, however, as a large model, its inference time exceeded 0.75 second, which cannot be acceptable for scenarios requiring real-time registration.
In contrast, our ModeTv2 had a faster speed with the help of custom CUDA operators.
Specifically, our ModeTv2-s used $0.23$s to predict one case, while our largest model ModeTv2-diff-l only required $0.30$s.
Note that our ModeTv2-s required much less memory usage than PCnet.

During pairwise optimization, when the network had to restart training, our small model took approximately $1.1$s for one round of training, while the large model took around $1.6$s.
This was faster compared to the majority of pyramid networks.
Compared to our previous ModeT, with CUDA operators, the training speed of ModeTv2-s was improved by around $7.1$ times, and the memory usage was only approximately $47\%$.
The testing speed of ModeTv2-s was improved by around $2.4$ times with CUDA operators, and the memory usage was only approximately $36\%$ of the original.
This shows the high efficiency of our CUDA operators.
It is worth noting that our ModeTv2-s can even be trained on a 2080Ti GPU with only 9GB of memory, similar to VM.

\begin{figure}[t]
	\centering
	\begin{overpic}[width=1\columnwidth]{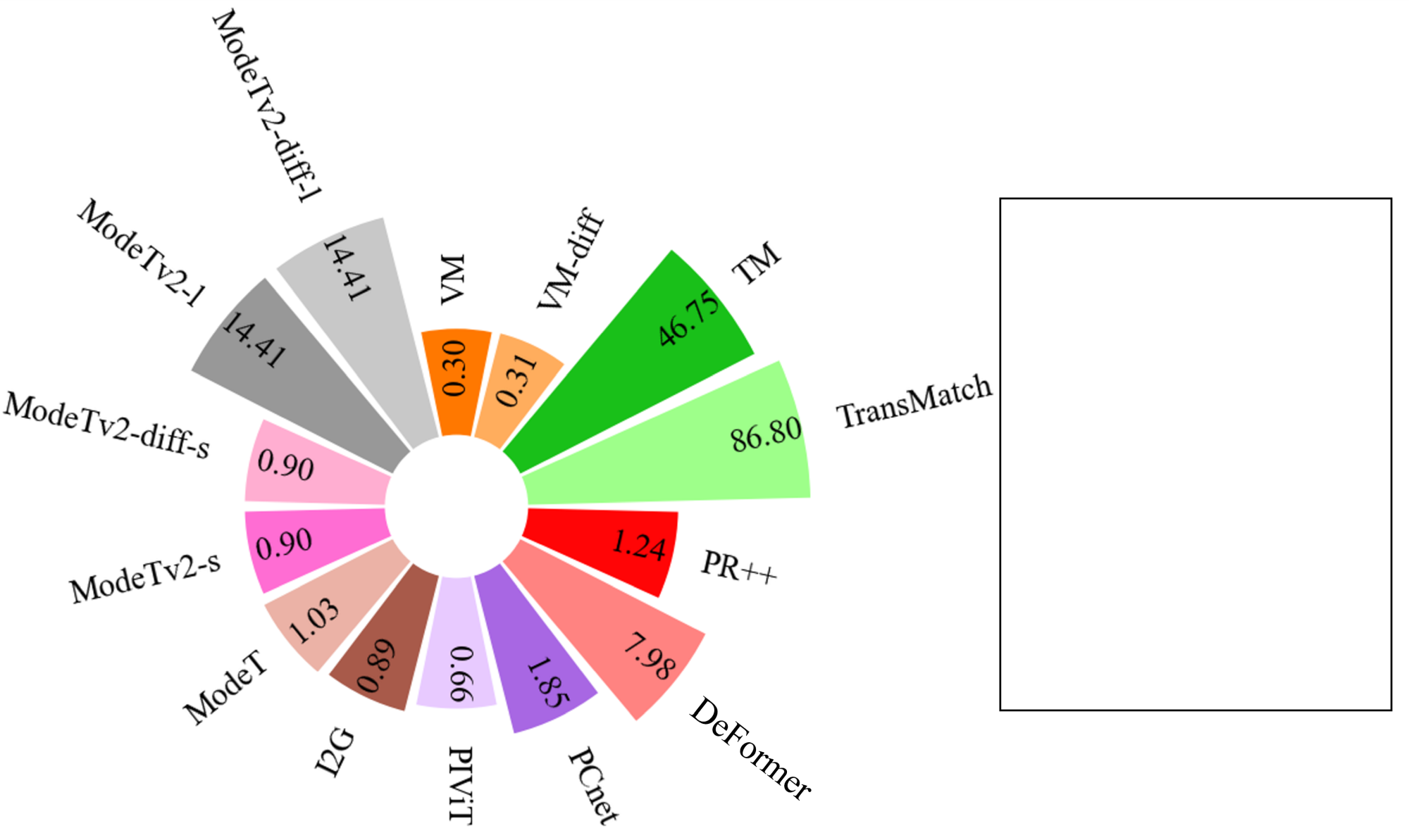}
		\put(71.5,42.5){\fontsize{5.5pt}{\baselineskip}\selectfont VM~\citep{VoxelMorph}}
		\put(71.5,40){\fontsize{5.5pt}{\baselineskip}\selectfont VM-diff~\citep{dalca2019unsupervised}}
		\put(71.5,37.5){\fontsize{5.5pt}{\baselineskip}\selectfont TM~\citep{Chen2022a}}
		\put(71.5,35){\fontsize{5.5pt}{\baselineskip}\selectfont TransMatch~\citep{chen2023transmatch}}
		\put(71.5,32.5){\fontsize{5.5pt}{\baselineskip}\selectfont PR++~\citep{Dual}}
		\put(71.5,30){\fontsize{5.5pt}{\baselineskip}\selectfont DeFormer~\citep{Chen2022}}
		\put(71.5,27.5){\fontsize{5.5pt}{\baselineskip}\selectfont PCnet~\citep{Lv2022}}
		\put(71.5,25){\fontsize{5.5pt}{\baselineskip}\selectfont PIViT~\citep{ma2023pivit}}
		\put(71.5,22.5){\fontsize{5.5pt}{\baselineskip}\selectfont I2G~\citep{Liu2022}}
		\put(71.5,20){\fontsize{5.5pt}{\baselineskip}\selectfont ModeT~\citep{wang2023modet}}
		\put(71.5,17.5){\fontsize{5.5pt}{\baselineskip}\selectfont ModeTv2-s}
		\put(71.5,15){\fontsize{5.5pt}{\baselineskip}\selectfont ModeTv2-diff-s}
		\put(71.5,12.5){\fontsize{5.5pt}{\baselineskip}\selectfont ModeTv2-l}
		\put(71.5,10){\fontsize{5.5pt}{\baselineskip}\selectfont ModeTv2-diff-l}										
	\end{overpic}
	\caption{Comparison of trainable parameters (M) of different methods.}
	\label{f:params}
\end{figure}

By observing Fig.~\ref{f:params}, it is evident that VM and VM-diff had the smallest parameter counts, while the Transformer-based TM and TransMatch largely increased the number of parameters.
Our proposed small ModeTv2 had a relatively smaller number of parameters,
approaching the magnitude of VM,
whereas the large ModeTv2's parameter count was still smaller than TM and TransMatch.

\section{Discussion}
\subsection{Discussion on Registration Accuracy}
According to Tables~\ref{tab:abct},~\ref{tab:lpba},~\ref{tab:mind} and~\ref{tab:ixi}, our ModeTv2 demonstrates a high level of accuracy in registration.
The results across all four datasets surpass those of our conference version of ModeT,
by +1.1\%, +0.9\%, +1.3\%, +0.8\% on ABCT, LPBA, Mindboggle, and IXI datasets, respectively.
This improvement is mainly attributed to the design of RegHead.
RegHead serves two purposes.
One is to integrate multiple motion patterns and reevaluate unreasonable neighborhood relationships.
Additionally, with only one convolutional layer, RegHead minimally transforms semantic information for multiple motion subfields, maintaining the explicit semantic generation of the ModeTv2 operator.
This facilitates transferability, allowing for fine-tuning of the encoder during PO without relearning the deformation field estimation.

\subsection{Discussion on Convergence Ability}
Examining the convergence results in Fig.~\ref{fig:converge} reveals that our network achieves high DSC scores in the first round, thanks to its strong inductive bias, a combination of the pyramid structure and the ModeTv2 module for explicit deformation field calculation.
Comparing the convergence rates in Fig.~\ref{fig:converge}, it is evident that pyramid networks generally converge quickly.
Our network's faster convergence, compared to other pyramid networks, is primarily due to the ModeTv2 module, which decomposes motion patterns without the need for extensive learning.
With the CUDA-accelerated implementation, our ModeTv2 is effective in facilitating easier learning and faster convergence.

\subsection{Discussion on PO Ability}
In the PO stage, our network achieves convergence in a few rounds.
According to Fig.~\ref{fig:POres} and Table~\ref{tab:spme},
our network achieves convergence in approximately 5 rounds for PO on different brain MRI datasets, taking about $5.5$s to $7.5$s with the support of CUDA operators.
In PO on different modalities and different organs, convergence is reached in about 20 rounds, approximately $22$s to $30$s.
Compared to traditional SyN method, which requires over $100$s for PO, our network demonstrates a speed advantage.

Examining the results, it is evident that the accuracy and generalization ability of the larger version of ModeTv2 are favorable.
The smaller version of ModeTv2 is more convenient for deployment and exhibits faster execution speed.
This presents a trade-off scenario, where the choice of model may need to be considered based on the specific application.
For instance, in scenarios requiring real-time surgical applications, the small model might be a preferable choice due to its efficiency.
On the other hand, situations with ample computational resources, less emphasis on real-time requirements, and a higher demand for accuracy might favor the deployment of the large model.

\subsection{Discussion on Interpretability}
Currently, deep learning methods for image registration lack sufficient interpretability.
In the case of same-domain tasks, enhancing the interpretability of DL registration models can improve the plausibility of the resulting deformation fields and accelerate model convergence.
Firstly, by enhancing interpretability, we enable the model to compute deformation fields in a pre-defined, reasonable manner.
This contrasts with other DL-based registration methods that only impose constraints on the output, thus enhancing the plausibility of the final deformation fields.
Secondly, increasing interpretability effectively strengthens the model's inductive bias for registration, which, in turn, accelerates convergence.

As we mentioned in the Introduction Section, we argue that adopting network designs closely aligned with the computational aspects of the registration task can alleviate the learning burden on the network and enhance convergence efficiency.
Some methods use cross-attention mechanisms to capture spatial relationships between moving and fixed image features, aiming to enhance the model's inductive bias for registration.
However, most current methods utilize this mechanism merely to enhance feature representation, with few considering its direct usage for solving the deformation field.
In contrast, our motion decomposition transformer simultaneously considers the decomposition of motion patterns in features and explicitly calculates the deformation field from the correlations of features, showcasing interpretability.

Regarding the plausibility of the deformation field, we observe from our accuracy experiments that the ModeT series achieve high accuracy while maintaining a low folding rate.
Even without using a diffeomorphic approach, the deformation fields generated by ModeTv2 are already quite plausible, and the deformation fields produced by ModeTv2-diff exhibit the lowest folding rate among DL methods.
Regarding convergence, our experimental results show that the ModeTv2 approaches convergence after just the first iteration.
This is due to our pre-defined approach for computing deformation fields in the decoder part of the network, which means that our model requires minimal learning on how to generate the deformation field.
Instead, it focuses more on learning how to extract image features, significantly improving convergence speed.

As for tasks in different domains, our interpretable model demonstrates strong generalization capabilities.
During the PO experiments, the network focuses primarily on feature extraction, while the interpretable deformation field computation requires little to no retraining.
The PO experiments have demonstrated the strong cross-domain ability of our model.

\section{Conclusion and Future Work}
We present a GPU-accelerated motion decomposition Transformer, namely ModeTv2, to naturally model the correspondence between images and convert it into deformation fields, which improves the interpretability and efficiency of the deep-learning-based registration networks.
The proposed ModeTv2 is built upon our MICCAI work ModeT.
It modifies the fusion method of multiple subfields from the original competitive weighting to the current RegHead.
This solves the problem that subfields in ModeT take the neighborhood relationship into account.
In order to provide efficient computing, which is an important issue for the volumetric registration, we design CUDA operators in ModeTv2.
Importantly, for different scenarios, we provide four versions of ModeTv2, i.e., the small model, large model, and their corresponding diffeomorphism models.
Experimental results demonstrate that the proposed method achieves superior performance over existing techniques in terms of both registration accuracy and efficiency.
We also conduct pairwise optimization experiments across different datasets.
By adopting the PO, our ModeTv2 balances accuracy, efficiency, and generalizability.

For future work, we envision several directions that could be further explored.
In this work, the encoder still adopts pure convolutions, however, Transformer-based encoders could be investigated, which may potentially reduce the overall trainable parameter count of the network.
Essentially, ModeTv2 uses attention coefficients to weight local deformation fields of different sizes.
Hence, more efficient attention mechanisms could be explored.
Moreover, applying the ModeTv2 for multi-modal registration tasks could be studied.
We hope this research can provide insights into the application of deep learning in the registration field.
We will further explore PO experiments in additional scenarios to extensively validate the robustness of our model.

\section*{Acknowledgments}
This work was supported in part by the National Natural Science Foundation of China under Grants 62471306, 62071305, 62171290 and 12326619,
in part by the Guangdong-Hong Kong Joint Funding for Technology and Innovation under Grant 2023A0505010021,
in part by the Shenzhen Medical Research Fund under Grant D2402010,
and in part by the Guangdong Basic and Applied Basic Research Foundation under Grant 2022A1515011241.

\bibliographystyle{model2-names.bst}\biboptions{authoryear}
\bibliography{bib2}


\end{document}